\documentclass[a4paper,fleqn]{cas-dc}

\usepackage[numbers]{natbib}
\usepackage{subfig}
\usepackage[ruled,longend,linesnumbered]{algorithm2e}
\def\tsc#1{\csdef{#1}{\textsc{\lowercase{#1}}\xspace}}
\tsc{WGM}
\tsc{QE}
\tsc{EP}
\tsc{PMS}
\tsc{BEC}
\tsc{DE}

\begin{document}
	\let\WriteBookmarks\relax
	\def\floatpagepagefraction{1}
	\def\textpagefraction{.001}
	\shorttitle{CATA++: A Collaborative Dual Attentive Autoencoder Method for Recommending Scientific Articles}
	\shortauthors{Meshal Alfarhood et~al.}
	
	\title [mode = title]{CATA++: A Collaborative Dual Attentive Autoencoder Method for Recommending Scientific Articles}

	\author[1]{Meshal Alfarhood}
	\cormark[1]
	\ead{may82@missouri.edu}
	\credit{Conceptualization, Methodology, Software, Validation, Writing - original draft}
	
	\author[1,2]{Jianlin Cheng}
	\ead{chengji@missouri.edu}
	\credit{Conceptualization, Supervision, Resources, Writing - review \& editing}
	
	\address[1]{Department of Electrical Engineering and Computer Science, University of Missouri, Columbia, MO 65211, USA}
	\address[2]{Informatics Institute, University of Missouri, Columbia, MO 65211, USA}
	
	\cortext[cor1]{Corresponding author}
	
	\begin{abstract}
		Recommender systems today have become an essential component of any commercial website. Collaborative filtering approaches, and Matrix Factorization (MF) techniques in particular, are widely used in recommender systems. However, the natural data sparsity problem limits their performance where users generally interact with very few items in the system. Consequently, multiple hybrid models were proposed recently to optimize MF performance by incorporating additional contextual information in its learning process. Although these models improve the recommendation quality, there are two primary aspects for further improvements: (1) multiple models focus only on some portion of the available contextual information and neglect other portions; (2) learning the feature space of the side contextual information needs to be further enhanced. 
		
		In this paper, we introduce a Collaborative Dual Attentive Autoencoder (CATA++) for recommending scientific articles. CATA++ utilizes an article's content and learns its latent space via two parallel autoencoders. We employ the attention mechanism to capture the most related parts of information in order to make more relevant recommendations. Extensive experiments on three real-world datasets have shown that our dual-way learning strategy has significantly improved the MF performance in comparison with other state-of-the-art MF-based models using various experimental evaluations. The source code of our methods is available at: https://github.com/jianlin-cheng/CATA.
	\end{abstract}
	
	
	
	
	

	\maketitle
	
	\section{Introduction}
	The amount of data created in the last few years is overwhelming. Interestingly, the data volume grows exponentially yearly compared to the years before, making the era of big data. This motivates and attracts researchers to utilize this massive data to develop more practical and accurate solutions in most computer science domains. For instance, recommender systems (RSs) are primarily a good solution to process massive data in order to extract useful information (e.g., users' preferences) to help users with personalized decision making.
	
	
	Scientific paper recommendations are very common applications for RSs. They are quite useful for scholars to be aware of related work in their research area. Generally, there are three common techniques for recommendations. First, collaborative filtering techniques (CF) are widely successful models that are applied to RSs. CF models depend typically on users’ ratings, such that users with similar past ratings are more likely to agree on similar items in the future. Matrix Factorization (MF) is one of the most popular CF techniques for many years due to its simplicity and effectiveness. MF has been widely used in the recommendation literature, such that many proposed models are enhanced versions of MF \cite{MF, PMF, BPMF, SPMF, MFimplicit}. However, CF models generally rely only on users' past ratings in their learning process, and do not consider other auxiliary data, which has been validated later to improve the quality of recommendations. For that reason, the performance of CF models decreases significantly when users have a limited amount of ratings data. This problem is also known as the data sparsity problem. 
	
	More recently, a lot of effort has been conducted to include item's information along with the user's ratings via topic modeling \cite{CTR, CTR-TR, HFT}. Collaborative Topic Regression (CTR) \cite{CTR} for example is composed of Probabilistic Matrix Factorization (PMF) and Latent Dirichlet Allocation (LDA) to utilize both user’s ratings and item's reviews to learn their latent features. By doing that, the natural sparsity problem could be alleviated, and these kinds of approaches are called hybrid models. 
	
	Simultaneously, deep learning (DL) has gained an increasing attention in the recent years due to how it enhances the way we process big data, and to its capability of modeling complicated data such as texts and images. Deep learning takes part in the research of recommendation systems the last few years and surpasses traditional collaborative filtering methods. Restricted Boltzmann Machines (RBM) \cite{RBM} is one of the first works that applies DL for CF recommendations. However, RBM is not deep enough to learn users' tastes from users' feedback data. Following that, Collaborative Deep Learning (CDL) \cite{CDL} has been a very popular DL-based model, which extends the CTR model \cite{CTR} by replacing the LDA topic model with a Stacked Denoising Autoencoder (SDAE). In addition, Deep Collaborative Filtering (DCF) \cite{DCF} is a similar work that uses a marginalized Denoising Autoencoder (mDA) with PMF. Lately, Collaborative Variational Autoencoder (CVAE) \cite{CVAE} introduces a Variational Autoencoder (VAE) to handle items' contents. CVAE is evaluated against CTR and CDL and the experimental results show that CVAE has better predictions over CTR and CDL. 
	
	However, existing recommendation models, such as CDL and CVAE, have two major limitations. First, they assume that all features of the model's inputs are equally the same in contributing to the final prediction. Second, they focus only on some parts of the auxiliary items data and neglect other parts, which can be also utilized in improving recommendations. 
	
	Consequently, we introduce a Collaborative Dual Attentive Autoencoder (CATA++) for recommending scientific papers. We integrate the attention technique into our deep feature learning procedure to learn from article's textual information (e.g., title, abstract, tags, and citations between papers) to enhance the recommendation quality. The features learned by each attentive autoencoder are employed then into the matrix factorization (MF) method for our final articles' suggestions. To show the effectiveness of our proposed model, we perform a comprehensive experiment on three real-world datasets to show how our model works compared to multiple recent MF-based models. The results show that our model can extract better features than other baseline models from the textual data. More importantly, CATA++ has a higher recommendation quality in the case of high, sparse data. 
	
	The main contributions of this work are summarized in the following points:
	\begin{itemize}
		\item We introduce CATA++, a Collaborative Dual Attentive Autoencoder, that has been evaluated on recommending scientific articles. We employ the attention technique into our model, such that only relevant parts of the data can contribute more in the item content's representation. This representation helps in finding similarities between articles.
		\item We exploit more article content into our deep feature learning process. To the best of our knowledge, our model is the first model that utilizes all article content including title, abstract, tags, and citations all together in one model by coupling two attentive autoencoder networks. The latent features learned by each network are then integrated into the matrix factorization method for our ultimate recommendations. 
		\item We evaluate our model using three real-world datasets. We compare the performance of our proposed model with five baselines. CATA++ achieves superior performance when the data sparsity is extremely high.
	\end{itemize} 
	
	The remainder of this paper is organized in the following order. First, we explain some essential preliminaries in Section 2. Second, our model, CATA++, is demonstrated in depth in Section 3. Third, we discuss the experimental results thoroughly in Section 4. Lastly, we conclude our work in Section 5. 
	
	\section{Preliminaries}
	The essential background to comprehend our model is explained in this section. We first describe the Matrix Factorization (MF) for implicit feedback problems. After that, we demonstrate the original idea of the attention mechanism and its related work.
	
	\subsection{Matrix Factorization}
	Matrix Factorization (MF) \cite{MF} is a very popular technique among CF models. The MF works by estimating the user-item matrix, \(R\in\mathbb{R}^{n\times m}\), using two matrices, \(U\in\mathbb{R}^{n\times d}\) and \(V\in\mathbb{R}^{m\times d}\), such that the dot product of \(U\) and \(V\) estimates the original matrix \(R\) as: $R \approx U \cdot V^T$. The value of $d$ corresponds to the dimension of the latent factors, such that $d \ll	min(n,m)$.
	
	\(U\) and \(V\) are optimized through minimizing the difference between the actual ratings and the predicted ratings, as the following:
	\begin{equation} \label{MF2}
	\mathcal{L} = \sum_{i,j\in R}\frac{I_{ij}}{2}(r_{ij} - u_i v_j^T)^2 + \frac{\lambda_u}{2} \left\Vert u_i\right\Vert^2 + \frac{\lambda_v}{2} \left\Vert v_j\right\Vert^2 
	\end{equation} 
	where \(I_{ij}\) is a variable that takes value 1 if user\(_i\) rates item\(_j\), and value 0 if otherwise. Also, \(||U||\) calculates the Euclidean norm and \(\lambda_u, \lambda_v\) are two regularization terms preventing the values of $U$ and $V$ from being too large. This avoid model overfitting.
	
	The previous objective function is designed for explicit data. However, explicit data (e.g., users' ratings) is not available all the time. As a result, implicit feedback (e.g., users' clicks) is utilized more frequently in recommendations. Thus, Weighted Regularized Matrix Factorization (WRMF) \cite{MFimplicit} modifies the objective function in Equation \ref{MF2} to make it work for implicit data, as the following:
	\begin{equation} \label{MF3}
	\mathcal{L} = \sum_{i,j\in R}\frac{c_{ij}}{2}(p_{ij} - u_i v_j^T)^2 + \frac{\lambda_u}{2} \left\Vert u_i\right\Vert^2 + \frac{\lambda_v}{2} \left\Vert v_j\right\Vert^2 
	\end{equation}
	where \(p_{ij}\) is the preference variable that takes value 1 if user\(_i\) interacts with item\(_j\), and value 0 if otherwise. Also, a confidence variable (\(c_{ij}\)) is given to each user-item pair for implicit data, such that \(c_{ij} = a\) when \(p_{ij} = 1\), and \(c_{ij} = b\) when \(p_{ij} = 0\), where \(a > b > 0\).
	
	\subsection{Attention mechanism}
	The idea of the attention mechanism is motivated by the human vision system and how our eyes pay attention and focus to a specific part of an image, or specific words in a sentence. In the same way, attention in deep learning can be described simply as a vector of weights to show the importance of the input elements. Thus, the intuition behind attention is that not all parts of the input are equally significant, i.e., only few parts are significant for the model. Attention was initially designed for image classification task \cite{AttentionFirst}, and then successfully applied in natural language processing (NLP) for machine translation task \cite{AttentionTranslate} when the input and the output may have different lengths. 
	
	Attention has also been successfully applied in different recommendation tasks \cite{attention, attention2, attention3, attention4, attention5, attention6}. For example, MPCN \cite{attention3} is a multi-pointer co-attention network that takes user and item reviews as input, and then extracts the most informative reviews that contribute more in predictions. Also, D-Attn \cite{attention5} uses a convolutional neural network with dual attention (local and global attention) to represent the user and the item latent representations similarly like matrix factorization approach. Moreover, NAIS \cite{attention6} employs attention network to distinguish items in a user profile, which have more influential effects in the model predictions.   
	\section{Methodology}
	\begin{figure*}
		\centering
		\includegraphics[width=0.75 \linewidth]{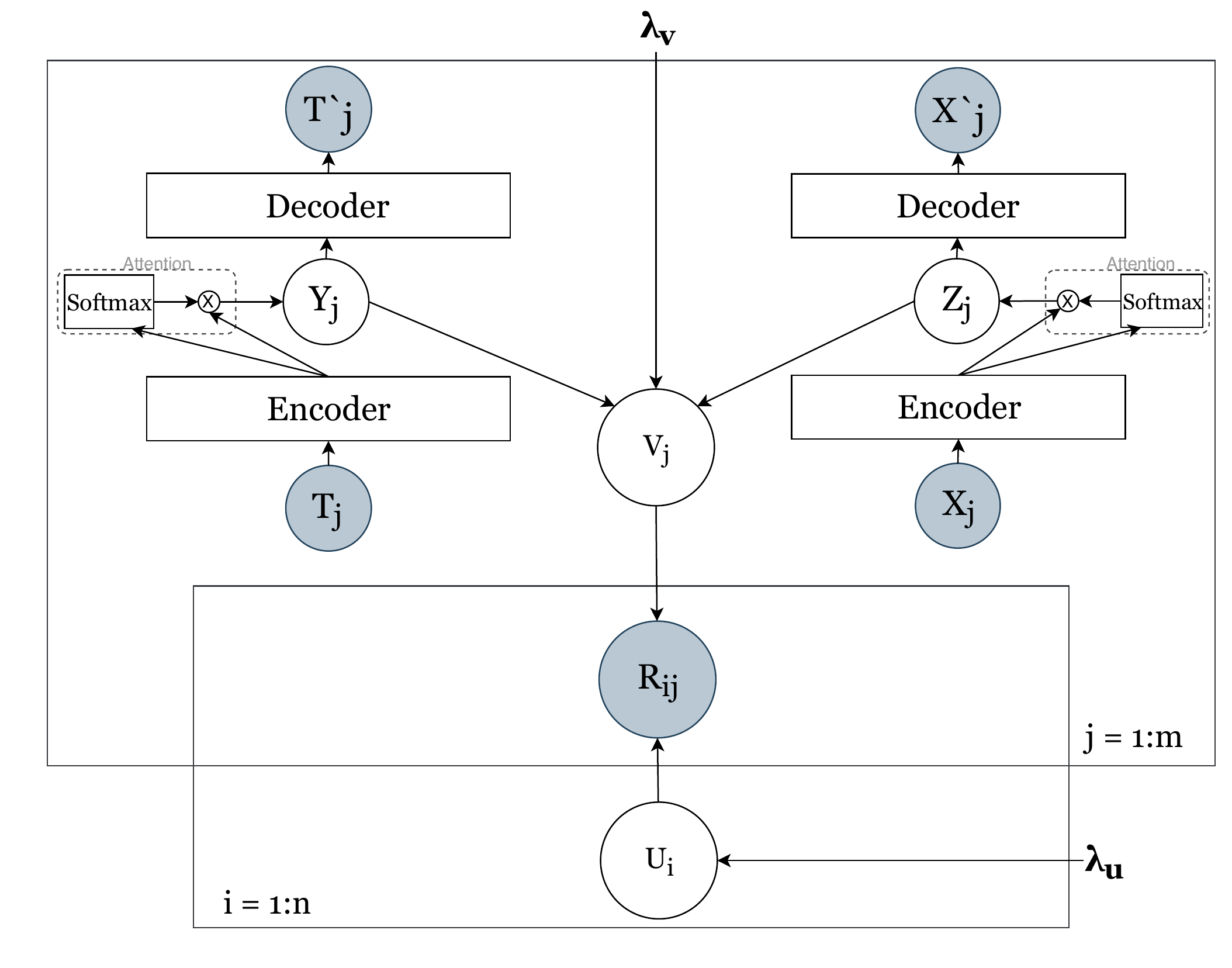}
		\caption{Collaborative Dual Attentive Autoencoder (CATA++) architecture.}
		\label{fig:CATA++}
	\end{figure*} 
	
	\begin{table}
		\centering
		\caption{A summary of notations used in this paper.}
		\label{notations}
		\begin{tabular}{c|l}
			\toprule
			\multicolumn{1}{c|}{Notation} &
			\multicolumn{1}{c}{Meaning}  \\
			\midrule
			n  & Number of users  \\
			m  & Number of articles   \\	
			d  & Dimension of the latent factors   \\	
			$R$  & User-article matrix \\	
			$U$  & Latent factors of users \\	
			$V$  & Latent factors of articles    \\	
			$C$  & Confidence matrix \\	
			$P$  & Users preferences matrix \\
			$X_j$ & Article's side information, i.e., title and abstract \\
			$T_j$ & Article's side information, i.e., tags and citations \\
			$\theta(X_j)$ & Mapping function for input $X_j$ of the autoencoder \\
			$\gamma(T_j)$ & Mapping function for input $T_j$ of the autoencoder \\
			$Z_j$ & Compressed representation of $X_j$  \\
			$Y_j$ & Compressed representation of $T_j$  \\
			$\lambda_u$ & Regularization parameter of users  \\
			$\lambda_v$ & Regularization parameter of articles  \\	
			\cmidrule(lr){1-2}
		\end{tabular}
	\end{table}
	
	Before describing our model thoroughly in this section, we first define the recommendation problem with implicit data and then followed by the illustration of our model.
	
	\subsection{Problem definition}
	The recommendation problem with implicit data is usually defined as the following:
	\begin{equation} \label{implicitFeedback}
	r_{ij}= 
	\begin{cases}
	1,		& \text{if $user_i$ interacts with $item_j$}\\
	0,      & \text{if otherwise}
	\end{cases}
	\end{equation} 
	where the ones refer to positive (observed) feedback, and the zeros refer to missing (unobserved) values. As a result, negative feedback is missing in this type of data. This problem is also called the one-class problem. There are multiple approaches have been proposed in regard to this issue. One popular solution is to sample negative feedback from the missing values. Another solution, which we adopt in this paper, is to assign different confidence values to all user-item pairs as we previously explained in the preliminaries section. Even though our model has been applied to a ranking predication problem with implicit feedback data, it could be used for a rating prediction problem with explicit feedback data as well by altering the final loss function.  
	
	In the following sections, we demonstrate each part of our model separately and we show how the recommendations are generated. Table \ref{notations} summarizes all the notations used in this paper and Figure \ref{fig:CATA++} displays the architecture of our model. 
	
	\subsection{The attentive autoencoder}
	The first part of our model is our deep, attentive autoencoder. Generally, an autoencoder \cite{AE} is a neural network that is trained in an unsupervised manner. Autoencoders are popular for dimensionality reduction, such that their input is compressed into low-dimensional representation while the features' abstract meaning is preserved. The autoencoder's network is composed of two main parts: the encoder and the decoder. The encoder takes an input and squashes it into a latent space, $Z_j$. The encoding function can be written as: \(Z_j=f(X_j)\). On the other hand, the decoder is used then to recreate the input again, $\hat{X_j}$, using the latent space representation ($Z_j$). The decoder function can be written as: \(\hat{X_j}=f(Z_j)\). Each of the encoder and the decoder consist usually of multiple hidden layers. The computations of the hidden layers are described as the following:  
	\begin{equation} \label{mlp}
	h^{(\ell)}  = \sigma (h^{(\ell-1)}W^{(\ell)} + b^{(\ell)})
	\end{equation}
	where $(\ell)$ points to the layer number, $W$ is the weights matrix, $b$ is the bias vector, and $\sigma$ is the Rectified Linear Unit (ReLU) activation function. 
	
	Our model takes two inputs from the article's data, $X_j=\{x^1,x^2,...,x^s\}$ and $T_j=\{t^1,t^2,...,t^g\}$, where $x^i$ and $t^i$ are real values between $[0,1]$, $s$ is the vocabulary size of the articles' titles and abstracts, and $g$ is the vocabulary size of the articles' tags. In other words, the inputs of our attentive network are two normalized bag-of-words histograms that represent the vocabularies of our articles' textual data.
	
	We apply the batch normalization (BN) \cite{BN} technique after each layer in our autoencoder to obtain a stable distribution of our output. Integrating BN into our model has an effective influence on our model accuracy; because it provides some regularization in training our neural network. Furthermore, we place an attention layer in the middle of our autoencoder, such that only the significant elements of the encoder's output are chosen to reconstruct the original input again. To do that, we use the \(softmax(.)\) function to compute the probability distribution of the encoder's output, as the following:
	\begin{equation} \label{softmax}
	f(z_c) = \frac{e^{z_c}}{\sum_{d} e^{z_d}} 
	\end{equation}
	
	After that, the output of the previous function and the encoder's output are multiplied by each other using the element-wise multiplication function to obtain the latent vector $Z_j$.
	
	Finally, we choose the binary cross-entropy as our objective function for each of our autoencoders, as the following:
	\begin{equation} \label{cross}
	\mathcal{L} = - \sum_{k} \big( y_k\log(p_k) -(1-y_k) \log(1-p_k) \big)
	\end{equation}
	where \(y_k\) refers to the correct values, and \(p_k\) refers to the predicted values. 
	
	The value of \(p\) that minimizes the previous loss function the most is when \(p=y\), which makes it fit for our autoencoder. To verify that, taking the derivative of the loss function with respect to p results in the following: 
	\begin{equation} \label{derivtives2} 
	\begin{aligned}
	&\frac{\partial \mathcal{L}}{\partial p} \ \  = -y(\frac{1}{p}) - (1-y)(\frac{-1}{1-p})  \\
	&\ \ \ \ \ \ \ \ \    \frac{-y}{p} + \frac{1-y}{1-p} = 0 \\
	&\ \ \ \ \ \ \ \ \    -y(1-p) + (1-y)p = 0\\
	&\ \ \ \ \ \ \ \ \    -y + yp + p - yp = 0\\
	&\ \ \ \ \ \ \ \ \    -y + p = 0\\
	&\ \ \ \ \ \ \ \ \ \   p = y
	\end{aligned} 
	\end{equation}

	\subsection{Probabilistic matrix factorization}
	Probabilistic Matrix Factorization (PMF) \cite{PMF} is a probabilistic linear model where the prior distributions of the users' preferences and the latent features are drawn from the Gaussian distribution. In our previous model, CATA \cite{CATA}, we train a single attentive autoencoder and incorporate its output into PMF. The objective function of CATA was defined as:
	\begin{equation} \label{CATA_loss}
	\mathcal{L} = \sum_{i,j\in R}\frac{c_{ij}}{2}(p_{ij} - u_i v_j^T)^2 + \frac{\lambda_u}{2} \left\Vert u_i\right\Vert^2 + \frac{\lambda_v}{2} \left\Vert v_j - \theta(X_j)\right\Vert^2 
	\end{equation} 
	where $\theta(X_j)=Encoder(X_j)=Z_j$ such that $\theta(X_j)$ works as the Gaussian prior information to $v_j$.

	On the other hand, CATA++ exploits extra item content and trains them via two separate, parallel attentive autoencoders. We use the output of the two separated networks all together to be the prior information of the items' latent factors from PMF. Therefore, the new objective function of CATA++ is modified to the following function:
	\begin{equation} \label{CATA+_loss}
	\mathcal{L} = \sum_{i,j\in R}\frac{c_{ij}}{2}(p_{ij} - u_i v_j^T)^2 + \frac{\lambda_u}{2} \left\Vert u_i\right\Vert^2 + \frac{\lambda_v}{2} \left\Vert v_j - (\theta(X_j) + \gamma(T_j))\right\Vert^2 
	\end{equation} 
	where $\gamma(T_j)=Encoder(T_j)=Y_j$.
	
	Taking the partial derivative of $\mathcal{L}$ with respect to \(u_i\) in Equation \ref{CATA+_loss} determines the values of users' latent vectors that minimize the the previous objective function the most, as follows:
	\begin{equation} \label{derivtives} 
	\begin{aligned}
	&\frac{\partial \mathcal{L}}{\partial u_i} \ \  = -\sum_{j}c_{ij}(p_{ij} - u_i v_j^T)v_j +  \lambda_u u_i \\
	&0 \ \ \ \ \ \   = -C_i(P_i - u_i V^T)V +  \lambda_u u_i \\
	&0 \ \ \ \ \ \   = -C_iVP_i + C_iVu_i V^T +  \lambda_u u_i \\
	&VC_iP_i  = u_iVC_iV^T + \lambda_u u_i \\
	&VC_iP_i = u_i(VC_iV^T + \lambda_u I) \\
	&u_i \ \ \ \ \  = VC_iP_i (VC_iV^T + \lambda_u I)^{-1} \\
	&u_i \ \ \ \ \  = (VC_iV^T+\lambda_uI)^{-1}VC_iP_i
	\end{aligned} 
	\end{equation}
	where \(I\) is the identity matrix.
	
	Similarly, taking the derivative of $\mathcal{L}$ with respect to \(v_j\) leads to: 
	\begin{equation} \label{UV} 
	\begin{aligned}
	v_j = (UC_jU^T+\lambda_vI)^{-1}UC_jP_j + \lambda_v (\theta(X_j) + \gamma(T_j))
	\end{aligned} 
	\end{equation}
	
	Finally, we use the Alternating Least Squares (ALS) optimization method to update the values of \(U\) and \(V\). It works by iteratively optimizing the values of $U$ while the values of $V$ are fixed, and vice versa. This operation is repeated until the values of \(U\) and \(V\) converge.
	
	\subsection{Prediction}
	Once we finish training our model, the model's prediction scores are computed as the dot product of the latent factors of users and articles (\(U\) and \(V\)). Specifically, each user's vector (\(u_i\)) is dot product with all vectors in \(V\), as: \(scores_i = u_i V^T\). As a result, we have a vector of different scores that represent the user's preferences. We then sort these scores in descending order, such that the top-\(K\) articles based on those scores are recommended. We repeat this process for all users. The overall process of our approach is illustrated in Algorithm \ref{alg:CATA++}. 
	
	\begin{algorithm}
		\SetAlgoLined
		\texttt{pre-train first autoencoder with input $X$\;}
		\texttt{pre-train second autoencoder with input $T$\;}
		\textbf{$Z$} $\gets \theta(X)$\;
		\textbf{$Y$} $\gets \gamma(T)$\;
		\texttt{\textbf{U, V} $\gets$ Initialize with random values\;}
		\While{\texttt{<NOT converge>}}{
			\For{\texttt{<each user $i$>}} {
				\texttt{\textbf{$u_i$} $\gets$ update using Equation \ref{derivtives}\;}
			}
			\For{\texttt{<each article $j$>}} {
				\texttt{\textbf{$v_i$} $\gets$ update using Equation \ref{UV}\;} 
			}
		}
		
		\For{\texttt{<each user $i$>}} {
			\texttt{\textbf{$scores_i$} $\gets u_i V^T$\;}
			\texttt{\textbf{sort($scores_i$)} in descending order\;}
		}
		\texttt{\textbf{Evaluate} the top-$K$ recommendations\;}
		\caption{CATA++ algorithm}
		\label{alg:CATA++}
	\end{algorithm}
	\section{Experiments}
	This section shows a comprehensive experiment in order to address the following research questions: 
	\begin{itemize}
		\item \textbf{RQ1}: How does our model perform compared to the state-of-the-art models? Prove with quantitative and qualitative analysis.
		\item \textbf{RQ2}: Are both autoencoders (left and right) cooperating with each other to enhance recommendation performance?  
		\item \textbf{RQ3}: What is the impact of different hyper-parameters tuning (e.g. dimension of features' latent space, number of layers inside each encoder and decoder, and regularization terms $\lambda_u$ and $\lambda_v$) on the performance of our model?
	\end{itemize} 
	Before answering the aforementioned research questions, we first describe the datasets, the evaluation metrics, and the baseline approaches against which we evaluate our model.
	
	\subsection{Datasets}
	\begin{table*}
		\centering
		\caption{Description of CiteULike datasets.}
		\label{dataset}
		\begin{tabular}{ccccccc}
			\toprule
			Dataset& \#Users& \#Articles & \#Pairs  & \#Tags & \#Citations & Sparsity\%\\
			\midrule
			Citeulike-a & 5,551  & 16,980 & 204,986  & 46,391 & 44,709 & 99.78\% \\
			Citeulike-t & 7,947 & 25,975 & 134,860 & 52,946 & 32,565 & 99.93\% \\
			Citeulike-2004-2007 & 3,039 & 210,137 & 284,960 & 75,721 & -- & 99.95\% \\
			\bottomrule
		\end{tabular} 
	\end{table*} 
	
	We use three real-world, scientific article datasets to evaluate our model against the state-of-the-art models. All three datasets are gathered from CiteULike website\footnote{www.citeulike.org}. CiteULike was a web service that let users to create their own library of academic publications. 
	
	First, Citeulike-a dataset, which is gathered by \cite{CTR}, has 5,551 users, 16,980 articles, 204,986 user-article interaction pairs, 46,391 tags, and 44,709 citations between articles. The tags are single-word keywords that are generated by CiteULike users when they add an article to their library. Citations between articles are taken from Google Scholar\footnote{https://scholar.google.com}. The data sparsity of this dataset is considerably high with only around 0.22\% of the user-article matrix having interactions. Users have at least 10 articles in their library.
	
	Second, Citeulike-t dataset, which is gathered by \cite{CTR-TR}, has 7,947 users, 25,975 articles, 134,860 user-article interaction pairs, 52,946 tags, and 32,565 citations between articles. This dataset is actually sparser than the first one, such that only 0.07\% of the user-article matrix having interactions. Users have at least three articles in their library.
	
	Third, Citeulike-2004-2007 dataset is three times bigger than the previous ones with regard to the user-article matrix. The data values in this dataset are extracted between 11-04-2004 and 12-31-2007. It is gathered by \cite{citeulike20042007} and it has 3,039 users, 210,137 articles, 284,960 user-article interaction pairs, and 75,721 tags. Also, it is worth pointing out that citations data is not available in this dataset. This dataset is even the sparsest dataset in this experiment with sparsity equal to 99.95\%. Users have at least 10 articles in their library. On average, they have 94 articles in their library and each article are added only to one user library. Also, this dataset poses a scalability challenge for recommender systems because of its size. More information about the datasets are shown in Table \ref{dataset}.
	
\begin{figure}
		\centering
		\includegraphics[width=0.99 \linewidth]{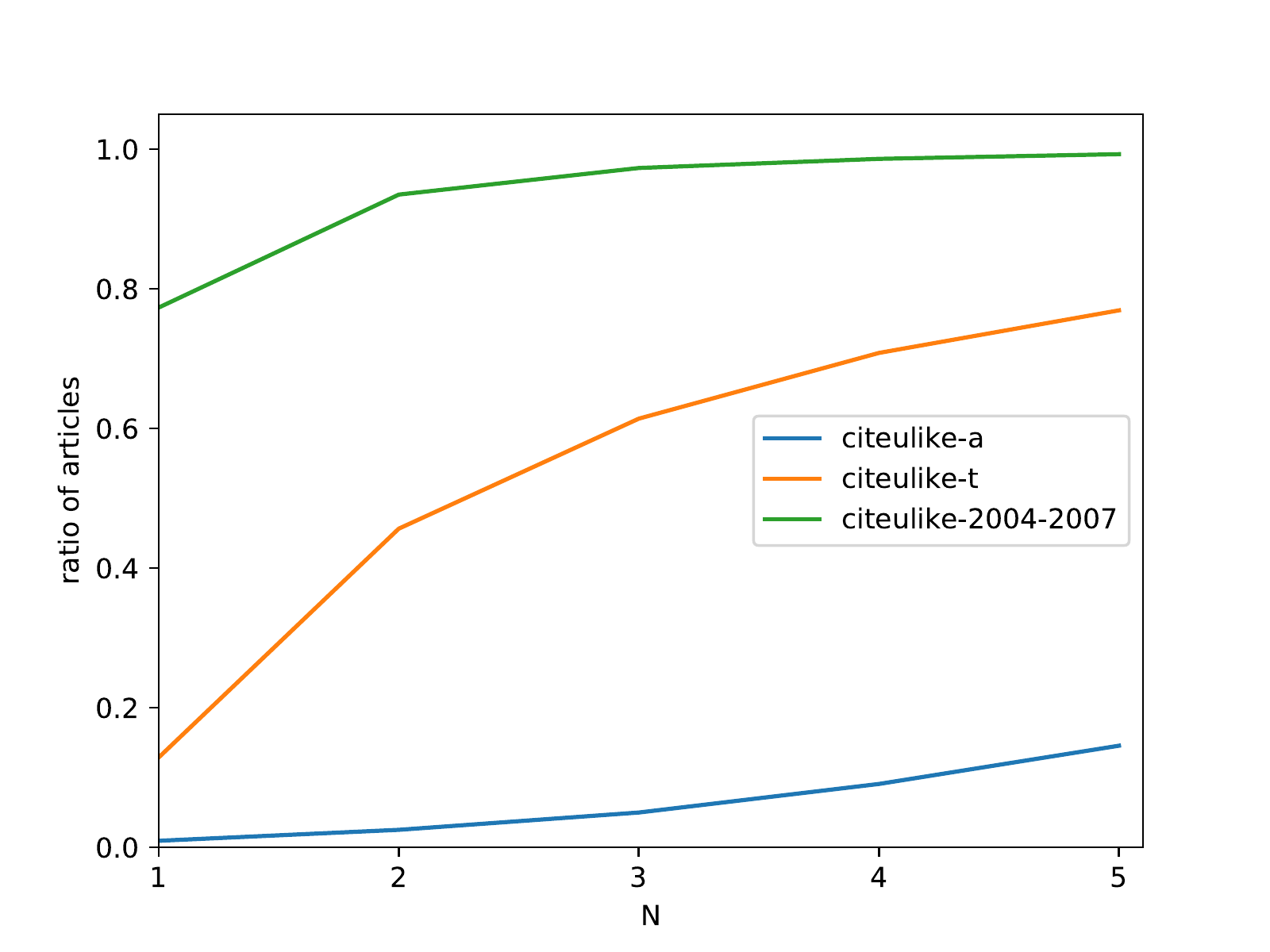}
		\caption{Ratio of articles that are added to $\leq$ N users' libraries.}
		\label{fig:ratio_items}
	\end{figure}

	Figure \ref{fig:ratio_items} shows the ratio of articles that are added to five or fewer users' libraries. For instance, 15\%, 77\%, and 99\% of the articles in Citeulike-a, Citeulike-t, and Citeulike-2004-2007, respectively, are added to five or fewer users' libraries. Moreover, only 1\% of the articles in Citeulike-a are added only to one user library, while the rest of the articles are added to more than this number. On the contrary, 13\%, and 77\% of the articles in Citeulike-t and Citeulike-2004-2007 are added only to one user library. This proves the sparseness of the data with regard to articles as we go from one dataset to another.
	
	We imitate the same procedure as the state-of-the-art models \cite{CDL, CTR, CVAE} to preprocess our textual data. First, we combine the title and the abstract of each article together. Second, we remove the stop words, such that the top-N unique words based on the TF-IDF measurement \cite{tfidf} are selected. As a result, 8,000, 20,000, and 19,871 words are selected for Citeulike-a, Citeulike-t, and Citeulike-2004-2007, respectively, to form the bag-of-words histograms. The bag-of-words histograms are normalized into values between zero and one based on the vocabularies' occurrences. The average number of words per article after our text preprocessing is 67, 19, and 55 words in Citeulike-a, Citeulike-t, and Citeulike-2004-2007, respectively.
	
	Similarly, we preprocess the tags information, such that tags assigned to fewer than five articles are removed, and thus we get 7,386 and 8,311 tags in total for Citeulike-a and Citeulike-t, respectively. For Citeulike-2004-2007 dataset, we only keep tags that are assigned to more than 10 articles, and that results in 11,754 tags in total for this dataset. After that, we create a matrix of bag-of-words histogram, \(Q\in\mathbb{R}^{m\times g}\), to represent the article-tag relationship, with \(m\) being the number of articles, and \(g\) being the number of tags. This matrix is filled with ones and zeros, such that:
	\begin{equation} \label{article-tag}
	q_{at}= 
	\begin{cases}
	1,		& \text{if tag$_t$ is assigned to article$_a$}\\
	0,      & \text{if otherwise}
	\end{cases}
	\end{equation}
	
	Also, citations between articles are integrated in this matrix, such that if \(article_x\) cites \(article_y\), then all the ones in vector \(q_y\) of the original matrix are copied into vector \(q_x\). We do that to capture the article-article relationship.
	
	\subsection{Evaluation methodology}
	To generate our training and testing data, we emulate the same procedure done in the state-of-the-art models ~\cite{CVAE, CDL, CTR-TR}. We generate the training and the testing data based on two settings, i.e., sparse and dense settings. To make the sparse (\(P=1\)) and the dense (\(P=10\)) datasets, we select \(P\) random articles from the user's library for the training data, while we select the remaining articles for the testing data. The data splitting is repeated four times. One split is used to run a validation experiment to fine-tune the hyper-parameters of each model, while the remaining three splits are utilized to report the average performance of each model.
	
	For our evaluation metrics, we adopt recall and normalized Discounted Cumulative Gain (nDCG). Recall per user is computed as the following:
	\begin{equation} \label{recall}
	recall@K=\frac{\text{Testing Articles} \cap \text{K Recommended Articles}}{\text{$\lvert$ Testing Articles $\rvert$}}
	\end{equation}
	
	However, the recall metric does measure the ranking quality within the top-\(K\) recommendations. Therefore, we use nDCG as well to show the ability of the model to recommend articles at the top of the ranking list. nDCG is computed as the following:
	
	\begin{equation} \label{ndcg}
	\begin{aligned}
	nDCG@K = \frac{1}{|U|}\sum_{u=1}^{|U|} \frac{DCG@K}{IDCG@K}
	\end{aligned}
	\end{equation}
	such that:
	\begin{equation} \label{dcg}
	\begin{aligned}
	&DCG@K=\sum_{i=1}^{K}\frac{\alpha(i)}{log_2(i+1)} \\
	&IDCG@K=\sum_{i=1}^{min(R,K)}\frac{1}{log_2(i+1)}
	\end{aligned}
	\end{equation}
	such that \(|U|\) refers to the number of users, \(i\) is the article rank, \(R\) is the number of relevant articles, and \(\alpha(i)\) is a variable that takes value 1 if the article is relevant, and 0 if otherwise. 
	
	\subsection{Baselines}
	\begin{table*} 
		\centering
		\caption{Comparison between all models about which data is used in their model training. }
		\label{contextual}
		\begin{tabular}{cccccc}
			\toprule
			\multicolumn{1}{c}{Approach} &
			\multicolumn{1}{c}{User-article matrix}  &
			\multicolumn{4}{c}{Side information} \\
			\cmidrule(lr){3-6}
			\multicolumn{1}{c}{} &
			\multicolumn{1}{c}{}  &
			\multicolumn{1}{c}{Title}   &
			\multicolumn{1}{c}{Abstract}  &
			\multicolumn{1}{c}{Tags}&
			\multicolumn{1}{c}{Citations}  \\
			\midrule
			POP    & \checkmark & -- & -- & -- & --  \\	
			CDL   & \checkmark & \checkmark &  \checkmark & --  & --  \\
			CVAE   & \checkmark & \checkmark &  \checkmark & --  & --  \\
			CVAE++ & \checkmark & \checkmark & \checkmark & \checkmark & \checkmark  \\	
			CATA   & \checkmark & \checkmark & \checkmark & -- & --  \\	
			CATA++ & \checkmark & \checkmark & \checkmark & \checkmark & \checkmark  \\	
			\cmidrule(lr){1-6}
		\end{tabular}
	\end{table*}
	We evaluate our approach against the following methods: 
	\begin{itemize}
		\item \textbf{POP}: Popular predictor is a non-personalized recommender system. It recommends the most popular articles in the training set to all users. It is widely used as the baseline for personalized recommender systems models.
		\item \textbf{CDL}: Collaborative Deep Learning (CDL) ~\cite{CDL} is a probabilistic model that jointly models both ratings data and items data using a stacked denoising autoencoder (SDAE) and a probabilistic matrix factorization (PMF).
		\item \textbf{CVAE}: Collaborative Variational Autoencoder (CVAE) ~\cite{CVAE} is a similar approach to CDL \cite{CDL}. However, it uses a variational autoencoder (VAE) instead of SDAE to incorporate item content into PMF.
		\item \textbf{CVAE++}: We modify the implementation of CVAE ~\cite{CVAE} to include two variational autoencoders to engage more side information into the model training, like what CATA++ does. As a result of adding another VAE into the model, we change the loss function accordingly such that the loss of the item latent variable becomes: $\mathcal{L}(v)$ = \(\lambda_v \sum_{j} \left\Vert v_j - (z_j + y_j)\right\Vert_2^2 \), where $z_j$ is the latent content variable of the first VAE, and $y_j$ is the latent content variable of the second VAE. 
		\item \textbf{CATA}: Collaborative Attentive Autoencoder (CATA) \cite{CATA} is our preliminary work that uses a single attentive autoencoder (AAE) to train article content, i.e., title and abstract. 
	\end{itemize}
	
	Table \ref{contextual} gives more clarifications about which part of the article's data is involved in each model training. As the table shows, only CATA++ and CVAE++ use all the available information for training their model.
	
	Table \ref{uvTable} also reports the best values of $\lambda_u$ and $\lambda_v$ for CDL, CVAE, CVAE++, CATA, and CATA++ based on the validation experiment. We use a grid search of the following values \{0.01, 0.1, 1, 10, 100\} to obtain the optimal values. Moreover, for CDL, we set $a$=1, $b$=0.01, $d$=50, $\lambda_n$=1000, and $\lambda_w$=0.0001. Also, we use a 2-layer SDAE network architecture that has a structure of "\#Vocabularies-200-50-200-\#Vocabularies" to run their code on our datasets. Similarly, for CVAE and CVAE++, we also set $a$=1, $b$=0.01, and $d$=50. A three-layer VAE network architecture, which is similar to the structure reported in their paper, is used with a structure equivalent to "\#Vocabularies-200-100-50-100-200-\#Vocabularies". Finally, for CATA and CATA++, we also set $a$=1, $b$=0.01, and $d$=50. A four-layer AAE network architecture in the form of "\#Vocabularies-400-200-100-50-100-200-400-\#Vocabularies" is used train our models.
	
	\begin{table*}
		\centering
		\caption{Parameter settings for $\lambda_u$ and $\lambda_v$ for CDL, CVAE, CVAE++, CATA, and CATA++ based on the validation experiment. }
		\label{uvTable}
		\begin{tabular}{ccccccccccccc}
			\toprule
			\multicolumn{1}{c}{Approach} &
			\multicolumn{4}{c}{Citeulike-a}  &
			\multicolumn{4}{c}{Citeulike-t} &
			\multicolumn{4}{c}{Citeulike-2004-2007}\\
			\cmidrule(lr){2-5} \cmidrule(lr){6-9} \cmidrule(lr){10-13} 
			\multicolumn{1}{c}{} &
			\multicolumn{2}{c}{Sparse}  &
			\multicolumn{2}{c}{Dense}   &
			\multicolumn{2}{c}{Sparse}  &
			\multicolumn{2}{c}{Dense}&
			\multicolumn{2}{c}{Sparse}  &
			\multicolumn{2}{c}{Dense} \\
			\cmidrule(lr){2-3} \cmidrule(lr){4-5} \cmidrule(lr){6-7} \cmidrule(lr){8-9} \cmidrule(lr){10-11} \cmidrule(lr){12-13}
			& $\lambda_u$ & $\lambda_v$ & $\lambda_u$ & $\lambda_v$ & $\lambda_u$ & $\lambda_v$ & $\lambda_u$ & $\lambda_v$ & $\lambda_u$ & $\lambda_v$ & $\lambda_u$ & $\lambda_v$\\
			\midrule
			CDL         & 0.01 & 10 & 0.01 & 10 &  0.01 & 10 &  0.01 & 10 &  0.01 & 10  &  0.01 & 10   \\
			CVAE         & 0.1 & 10 & 1 & 10 & 0.1 & 10 & 0.1 & 10 & 0.1 & 10 & 0.1 & 10  \\
			CVAE++         & 0.1 & 10 & 0.1 & 10 & 0.1 & 10 & 0.1 & 10  & 1 & 10  & 1 & 10   \\	
			CATA      & 10 & 0.1 &  10 & 0.1  & 10 & 0.1 & 10 & 0.1 & 10 & 0.1 & 10 & 0.1  \\	
			CATA++      & 10 & 0.1 &  10 & 0.1  & 10 & 0.1 & 10 & 0.1 & 10 & 0.1 & 10 & 0.1  \\
			\cmidrule(lr){1-13}
		\end{tabular}
	\end{table*}
	\subsection{Experimental results}
	In this section, we address the research questions that are previously outlined at the start of the experiments section.
	
	\subsubsection{\textbf{RQ1}}
	\begin{figure*}
		\centering
		\subfloat[Citeulike-a]{\includegraphics[width=0.33\textwidth]{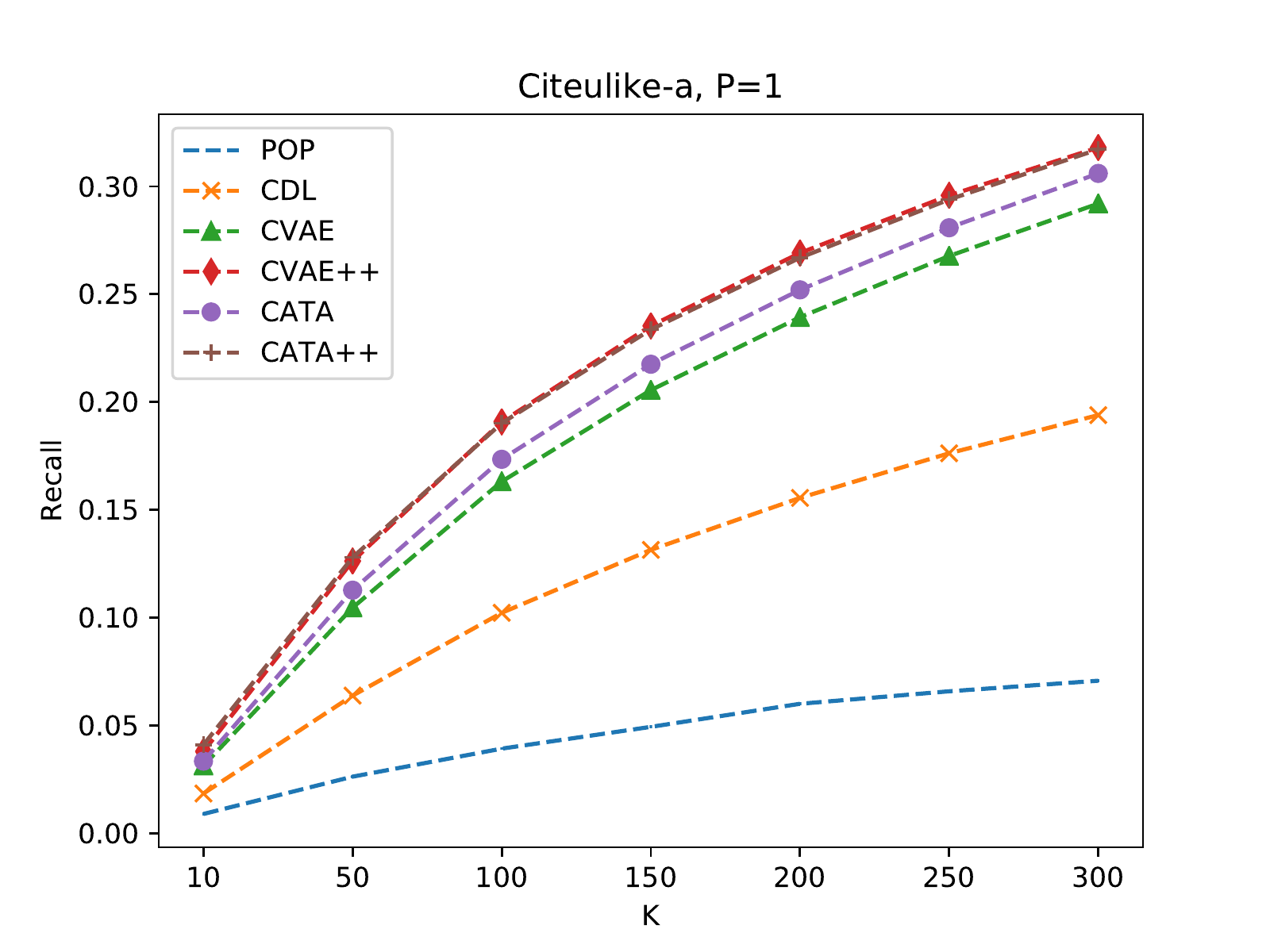}}
		\hfill
		\subfloat[Citeulike-t]{\includegraphics[width=0.33\textwidth]{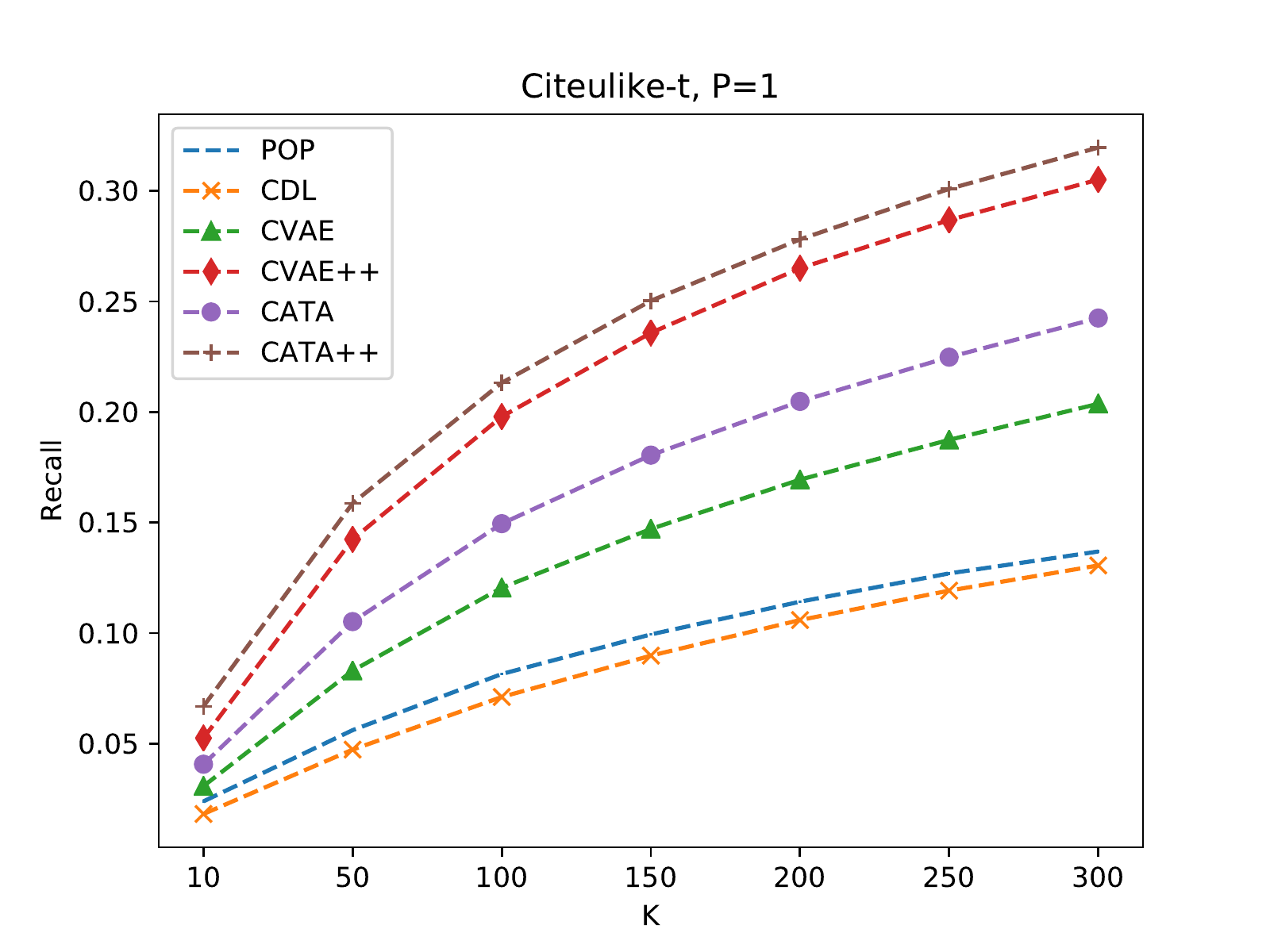}\label{fig:t1_recall}}
		\hfill
		\subfloat[Citeulike-2004-2007]{\includegraphics[width=0.33\textwidth]{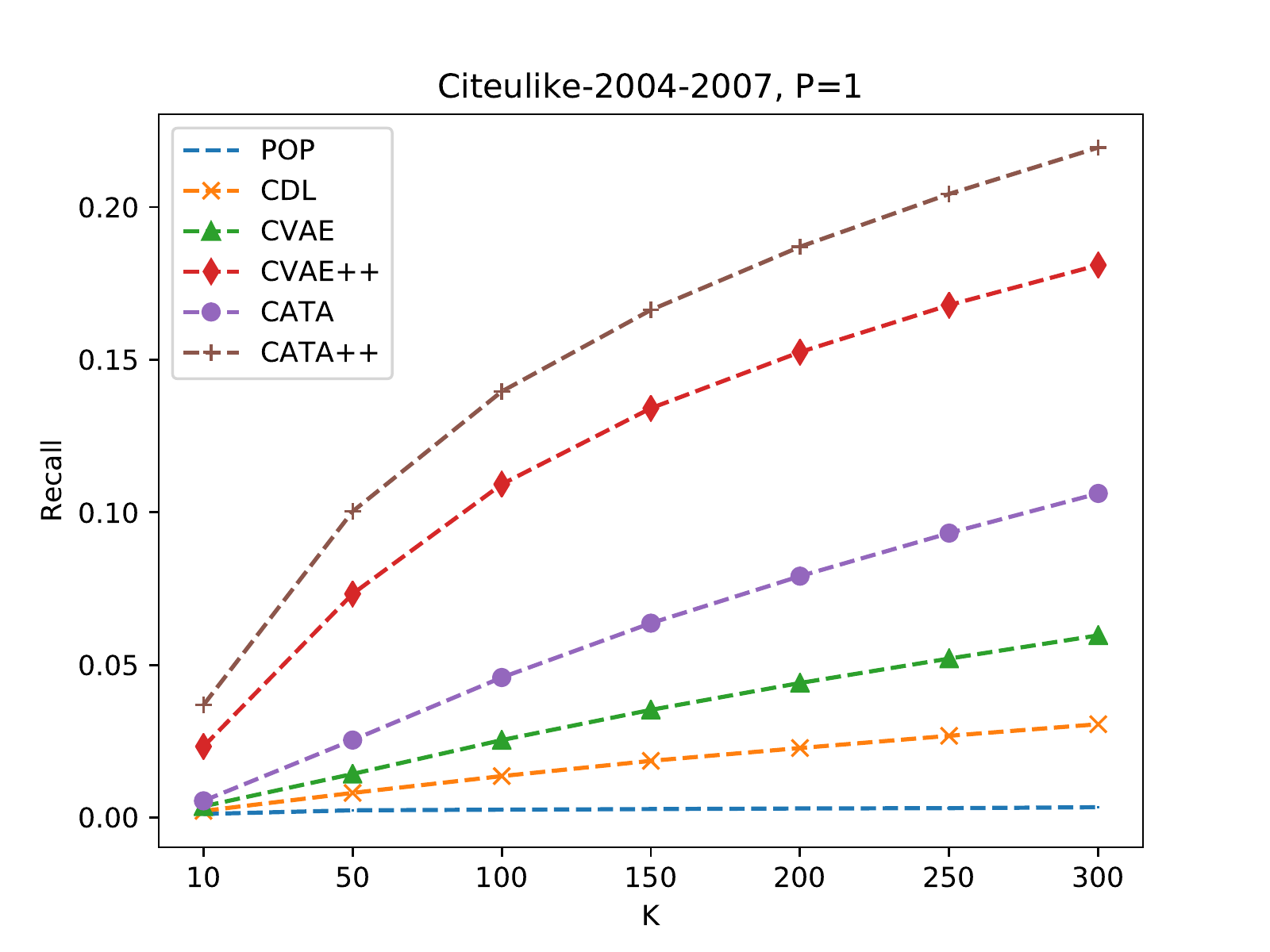}}
		\caption{The top-$K$ recommendation performance based on the recall metric using the sparse cases for (a) Citeulike-a, (b) Citeulike-t, and (c) Citeulike-2004-2007 datasets.}
		\label{fig:sparse_recall}
	\end{figure*}
	\begin{figure*}
		\centering
		\subfloat[Citeulike-a]{\includegraphics[width=0.33\textwidth]{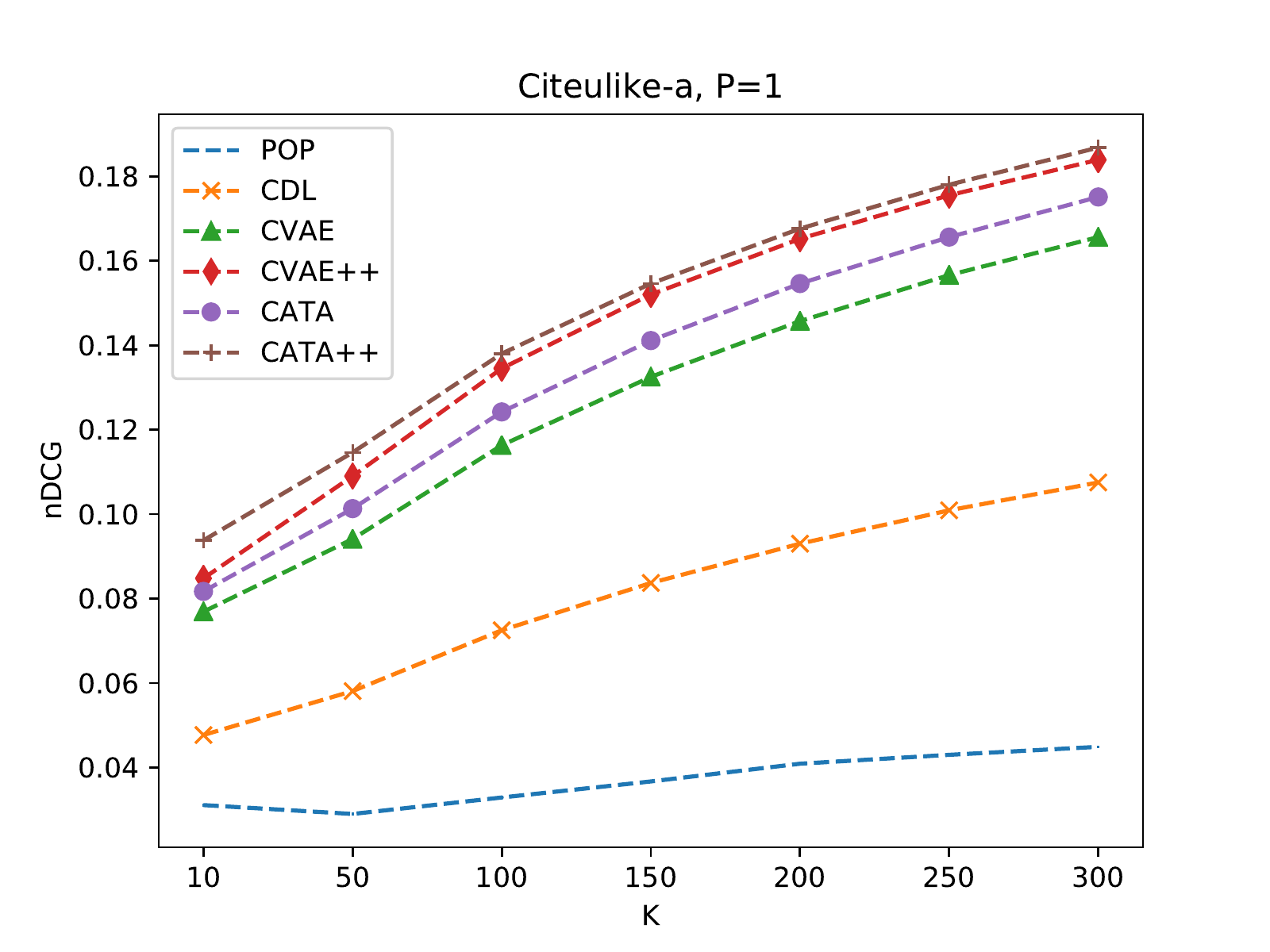}}
		\hfill
		\subfloat[Citeulike-t]{\includegraphics[width=0.33\textwidth]{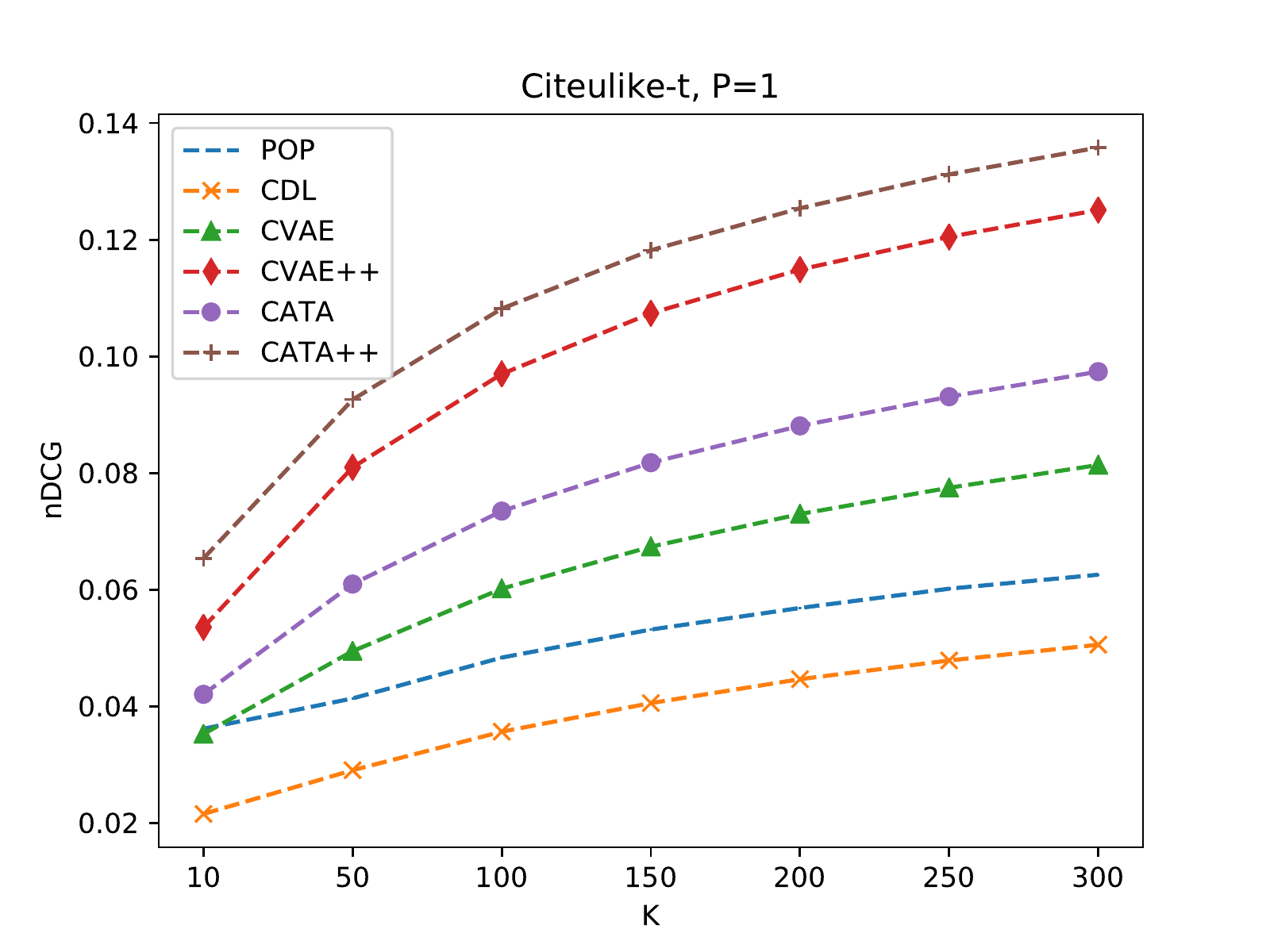}}
		\hfill
		\subfloat[Citeulike-2004-2007]{\includegraphics[width=0.33\textwidth]{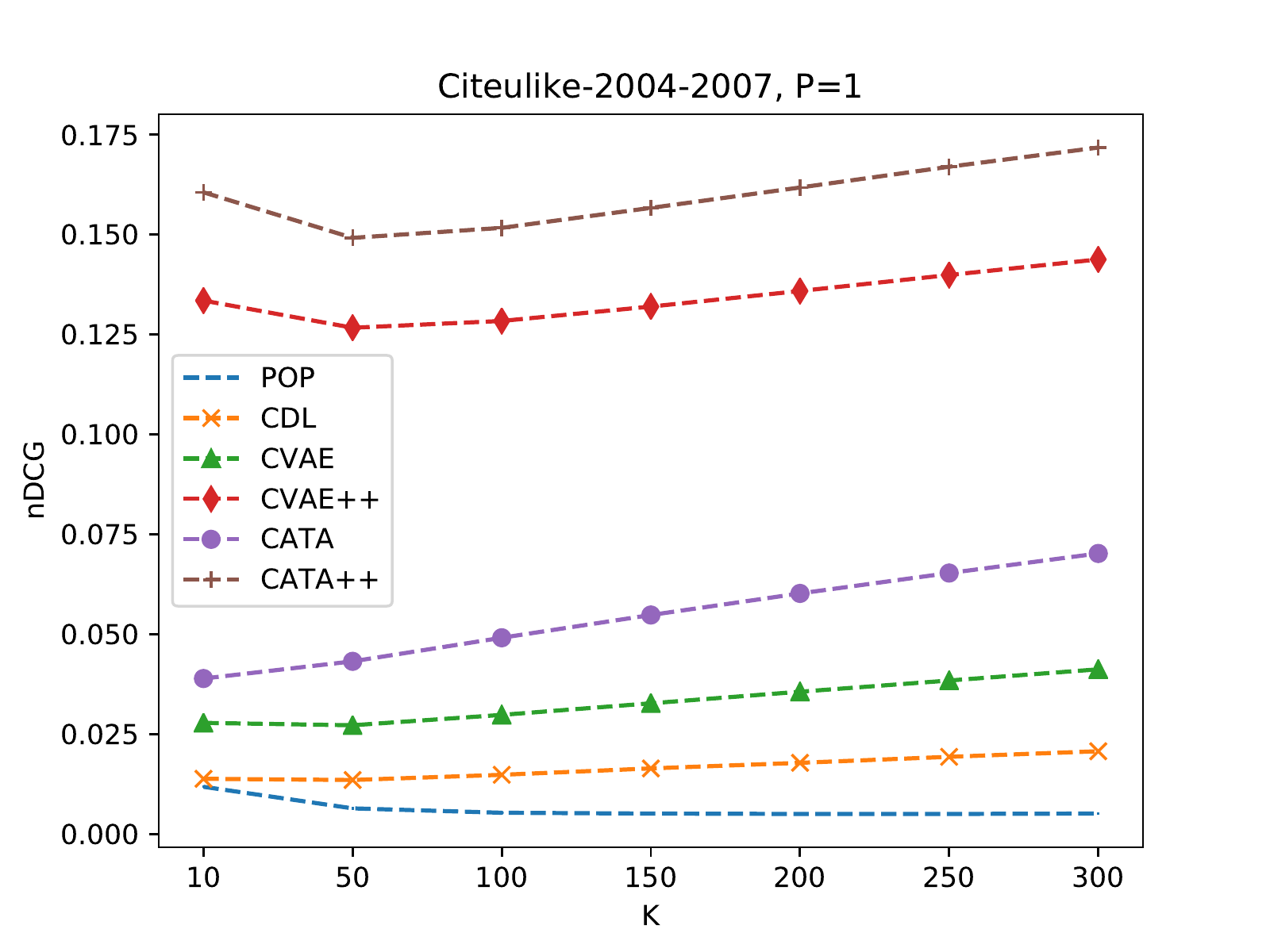}}
		\caption{The top-$K$ recommendation performance based on the nDCG metric using the sparse cases for (a) Citeulike-a, (b) Citeulike-t, and (c) Citeulike-2004-2007 datasets.}
		\label{fig:sparse_ndcg}
	\end{figure*}
	\begin{figure*}
		\centering
		\subfloat[Citeulike-a]{\includegraphics[width=0.33\textwidth]{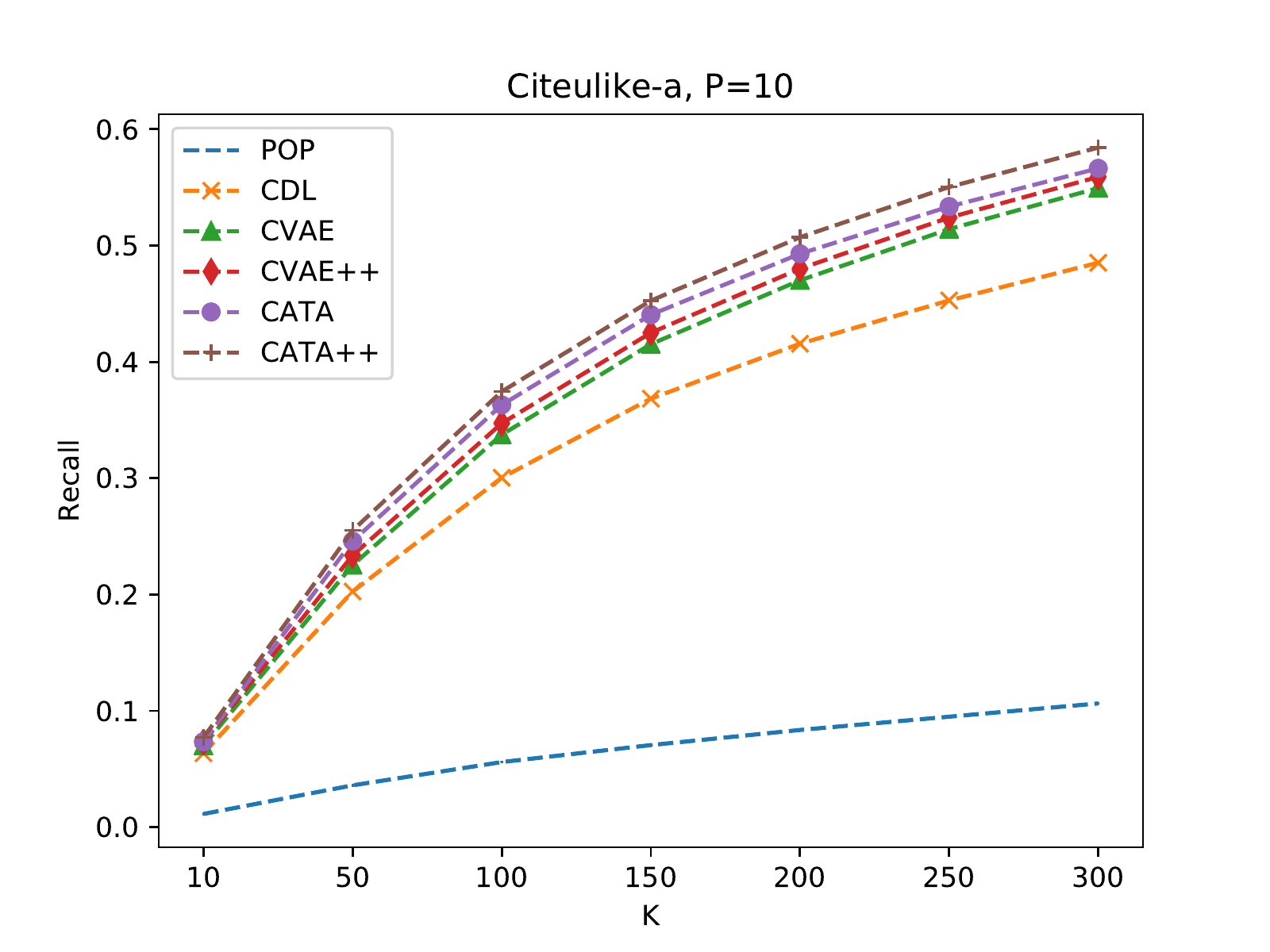}}
		\hfill
		\subfloat[Citeulike-t]{\includegraphics[width=0.33\textwidth]{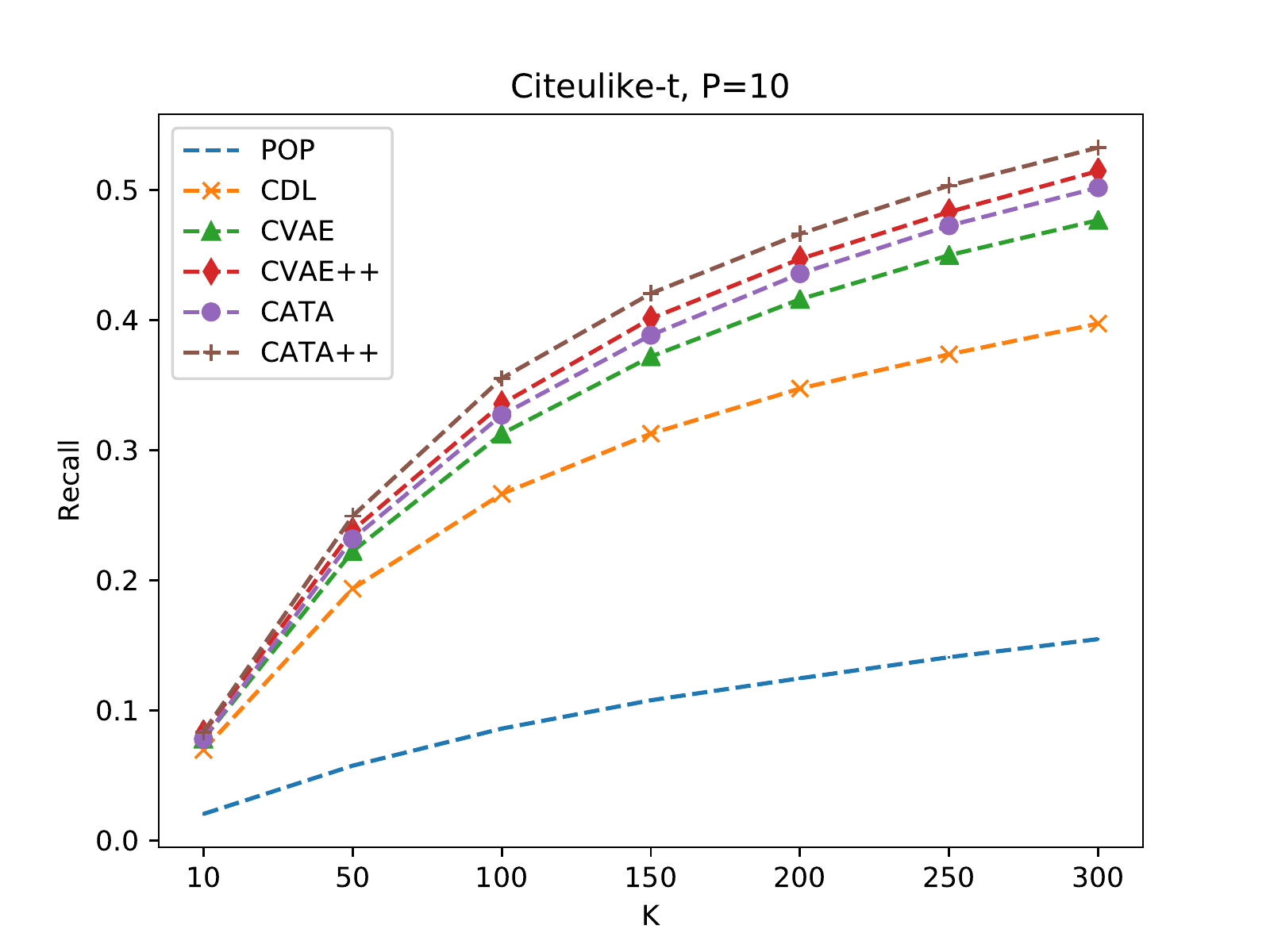}\label{fig:t10_recall}}
		\hfill
		\subfloat[Citeulike-2004-2007]{\includegraphics[width=0.33\textwidth]{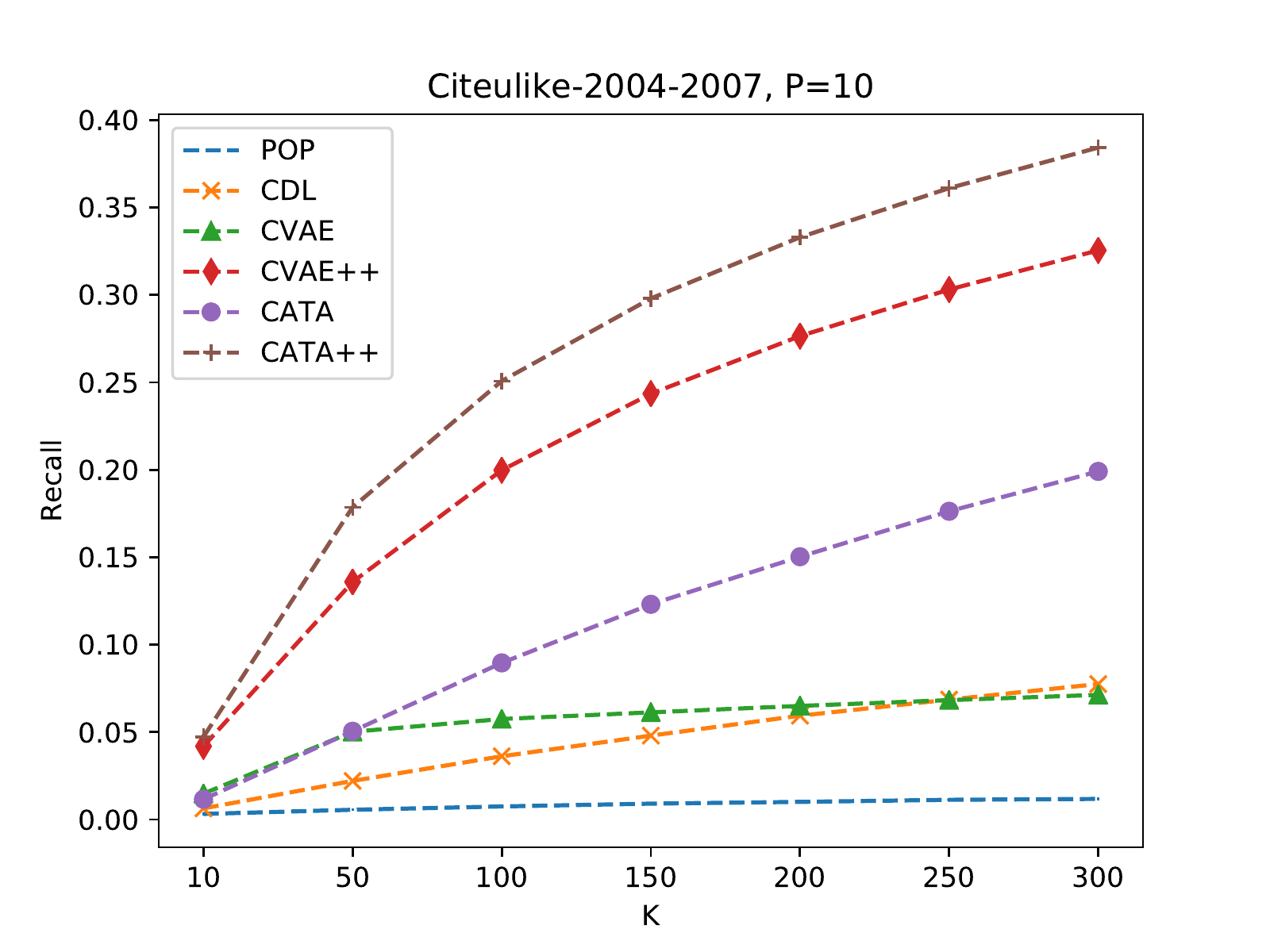}}
		\caption{The top-$K$ recommendation performance based on the recall metric using the dense cases for (a) Citeulike-a, (b) Citeulike-t, and (c) Citeulike-2004-2007 datasets.}
		\label{fig:dense_recall}
	\end{figure*}
	\begin{figure*}
		\centering
		\subfloat[Citeulike-a]{\includegraphics[width=0.33\textwidth]{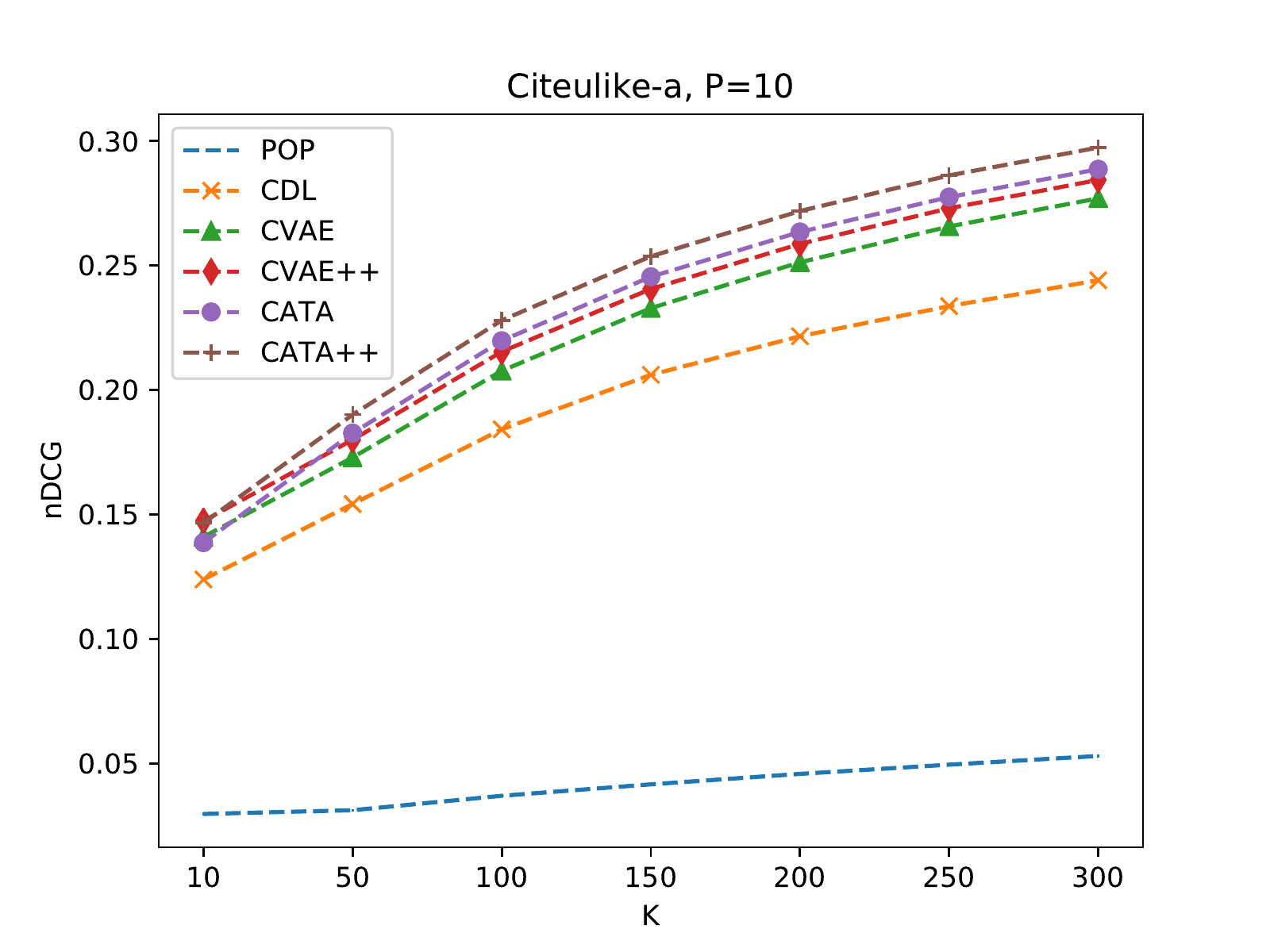}}
		\hfill
		\subfloat[Citeulike-t]{\includegraphics[width=0.33\textwidth]{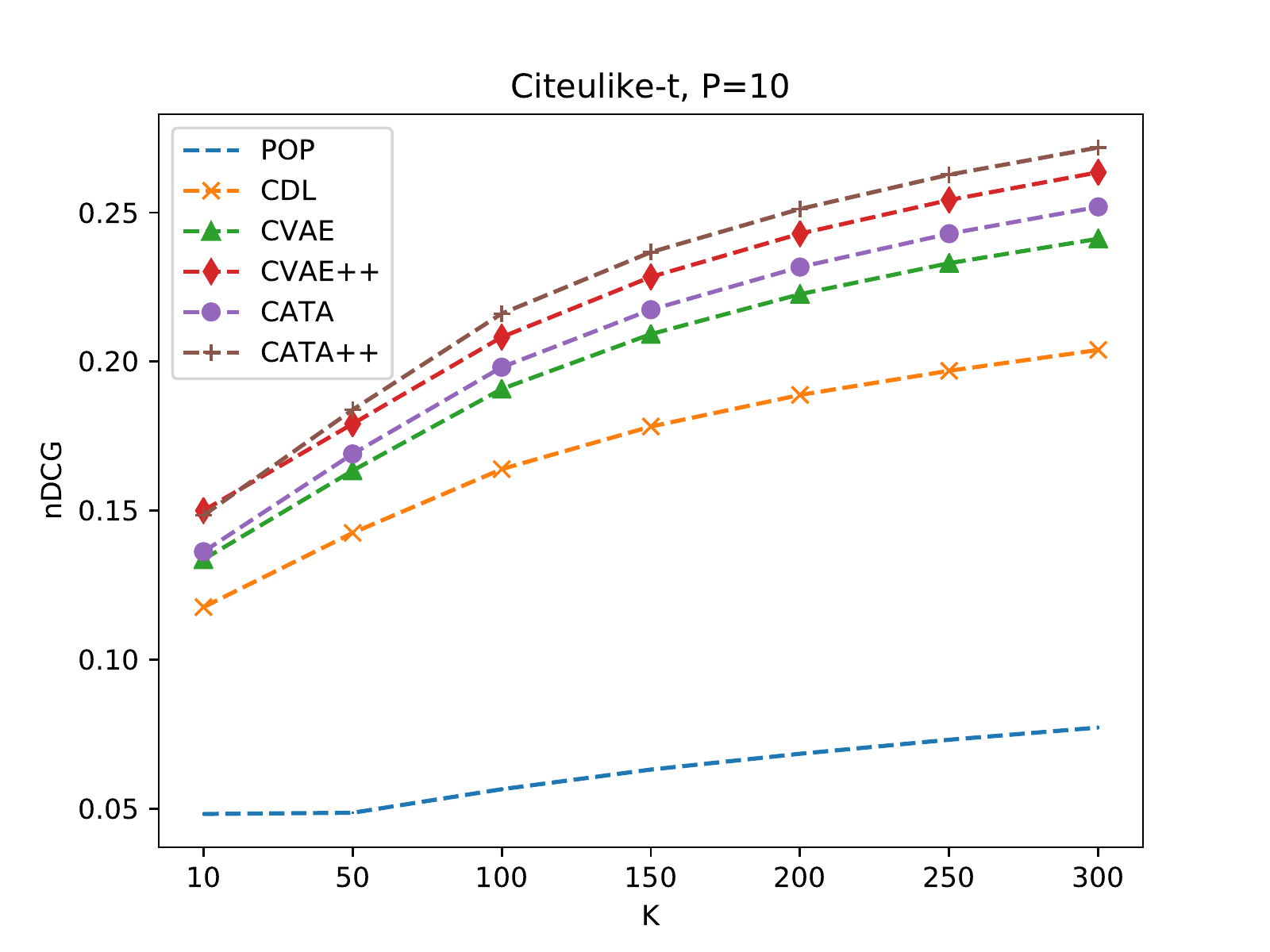}}
		\hfill
		\subfloat[Citeulike-2004-2007]{\includegraphics[width=0.33\textwidth]{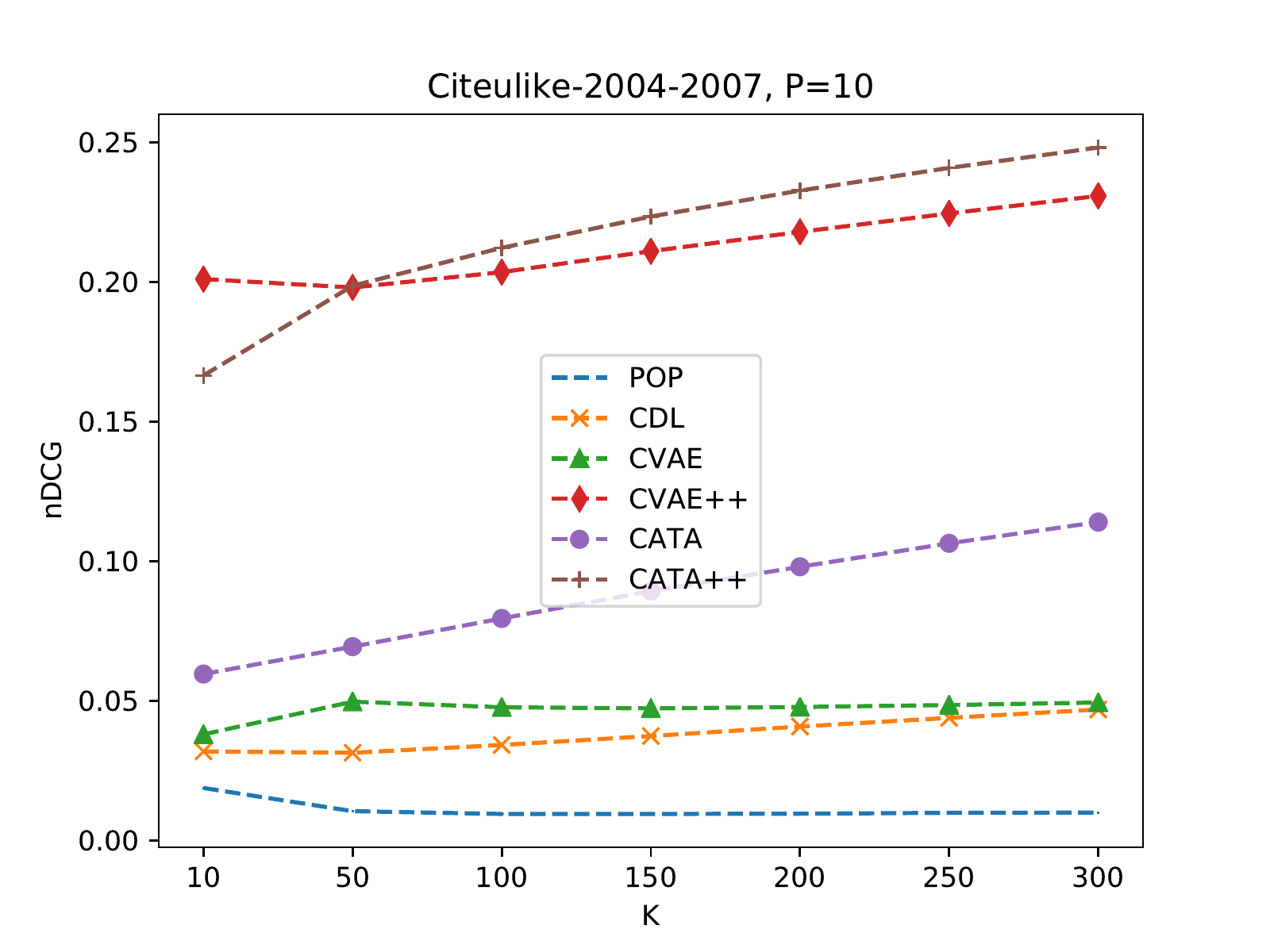}}
		\caption{The top-$K$ recommendation performance based on the nDCG metric using the dense cases for (a) Citeulike-a, (b) Citeulike-t, and (c) Citeulike-2004-2007 datasets.}
		\label{fig:dense_ndcg}
	\end{figure*}
	\begin{table*}
		\centering
		\caption{The improvement percentage of our model's performance over the best competitor according to Recall@10, Recall@300, nDCG@10, and nDCG@300. }
		\label{Improvement}
		\begin{tabular}{ccccc|cccc}
			\toprule
			\multicolumn{1}{c}{Approach} &
			\multicolumn{4}{c|}{Sparse}  &
			\multicolumn{4}{c}{Dense} \\
			\cmidrule(lr){2-5} \cmidrule(lr){6-9} 
			& Recall@10 & Recall@300 & nDCG@10 & nDCG@300 & Recall@10 & Recall@300 & nDCG@10 & nDCG@300 \\
			\midrule
			Citeulike-a          & 8.18\% & -- & 10.61\% & 1.57\%  & 4.89\%  & 3.16\% & -- & 3.01\%   \\	
			Citeulike-t         & 27.42\% & 4.75\% & 22.01\%  & 8.55\% & 0.84\% & 3.49\% & -- & 3.15\% \\
			Citeulike-2004-2007  & 58.36\% & 21.27\% & 20.29\%  & 19.47\%  & 12.88\% & 18.06\% & -- & 7.49\%   \\
			\cmidrule(lr){1-9}
		\end{tabular}
	\end{table*}
	
	\begin{table*}
		\centering
		\caption{A quality example of the top-10 recommendations using the sparse case of Citeulike-2004-2007 dataset.}
		\label{qualitative}
		\begin{tabular}{l|c}
			\toprule
			\multicolumn{2}{l}{User ID: 2214}  \\
			\cmidrule(lr){1-2}
			\multicolumn{2}{l}{Articles in training set: A collaborative filtering framework based on fuzzy association rules and multiple-level similarity}  \\
			\cmidrule(lr){1-2} 
			\multicolumn{1}{c|}{\textit{CATA++}} & In user library? \\
			\cmidrule(lr){1-2} 
			1. Item-based collaborative filtering recommendation algorithms & No \\
			2. Combining collaborative filtering with personal agents for better recommendations  & No \\
			3. An accurate and scalable collaborative recommender &  No \\
			\textbf{4. Google news personalization: scalable online collaborative filtering } &  \textbf{Yes} \\
			\textbf{5. Combining collaborative and content-based filtering using conceptual graphs } &  \textbf{Yes} \\
			6. Link prediction approach to collaborative filtering  &  No \\
			7. Slope one predictors for online rating-based collaborative filtering & No \\ 
			\textbf{8. Slope one predictors for online rating-based collaborative filtering} & \textbf{Yes} \\
			9. A decentralized CF approach based on cooperative agents  &  No \\
			\textbf{10. Toward the next generation of recommender systems: A survey of the state-of-the-art and possible...} & \textbf{Yes} \\
			\cmidrule(lr){1-2} 
			\multicolumn{1}{c|}{\textit{CVAE++}} & In user library? \\
			\cmidrule(lr){1-2} 
			1. Combining collaborative filtering with personal agents for better recommendations & No \\
			2. Explaining collaborative filtering recommendations & No \\
			\textbf{3. Google news personalization: scalable online collaborative filtering } & \textbf{Yes}\\
			4. Learning user interaction models for predicting web search result preferences & No \\
			5. Item-based collaborative filtering recommendation algorithms & No\\
			6. Enhancing digital libraries with TechLens+ & No\\  
			7. Optimizing search engines using clickthrough data & No \\
			8. Context-sensitive information retrieval using implicit feedback & No \\
			9. A new approach for combining content-based and collaborative filters & No \\
			\textbf{10. Combining collaborative and content-based filtering using conceptual graphs  } & \textbf{Yes}\\
			\midrule
		\end{tabular}
	\end{table*}
	To show how our model performs, we run quantitative and qualitative comparisons to address this matter. Figures \ref{fig:sparse_recall} and \ref{fig:sparse_ndcg} display the performance of the top-$K$ recommendations using the sparse data in terms of recall and nDCG for all the three datasets. Similarly, Figures \ref{fig:dense_recall} and \ref{fig:dense_ndcg} display the performance using the dense data in terms of recall and nDCG as well. 
	
	First, the sparse cases in Figures \ref{fig:sparse_recall} and \ref{fig:sparse_ndcg} are more challenging for any recommendation model; due to the scarce feedback data for the model's training. For the sparse cases, CATA++ achieves a superior performance relative to other MF-based models in all datasets based on the both metrics. More importantly, CATA++ beats the best model among all the baselines, CVAE++, by a wide margin in Citeulike-2004-2007 dataset, where it is actually sparser and contains a huge number of articles. This proves the validity of our model to work with sparse data. Second, for the dense cases, CATA++ again beats the other models as Figures \ref{fig:dense_recall} and \ref{fig:dense_ndcg} display. In reality, several of the existed methods actually perform well under this setting, but poorly when the sparsity is high. For example, CDL fails to beat POP in Citeulike-t dataset under the sparse setting, and then easily beats POP  under the dense setting as Figures \ref{fig:t1_recall} and \ref{fig:t10_recall} show.
	
	Consequently, this experiment validates the ability of our model to overcome the limitations mentioned in the beginning of this paper. For instance, among all the five baseline models, CVAE++ has the best performance, which emphasizes the usefulness of involving more article's data. Also, the attentive autoencoder (AAE) can extract more constructive information over the variational autoencoder (VAE) and the stacked denoising autoencoder (SDAE) as CATA has the superiority over CVAE and CDL, and CATA++ has the superiority over CVAE++ while utilizing the same data.
	
	Table \ref{Improvement} shows the percentage of performance improvement of our model, CATA++, over the best competitor among all baselines. This percentage measures the increase in performance, which can be calculated according to the following formula: $improv\% = (p_{our} - p_{sota}) / p_{sota} \times 100$, where $p_{our}$ is the performance of our model, and $p_{sota}$ is the performance of the best model among all baselines. 
	
	In addition to the aforementioned quantitative comparisons, qualitative comparisons are also reported in Table \ref{qualitative} to show the quality of recommendations using real examples. The table shows the top-10 recommendations generated by our model, CATA++, and the other competitive model, CVAE++, for one selected random user using Citeulike-2004-2007 dataset under the sparse setting. With this case study, we seek to gain a deeper insight into the difference between the two models in recommendations. The example in the table presents $user2214$ has only one article in his or her training library entitled "\textit{A collaborative filtering framework based on fuzzy association rules and multiple-level similarity}". This example defines the sparsity problem very well where a user has limited feedback data. Based on the article's title, this user is probably interested in recommender systems and more specifically in collaborative filtering (CF). After analyzing the results of each model, we can derive that our model can recommend more relevant articles than the other baseline. For instance, most of the top 10 recommendations based on CATA++ are related to the user's interest. The accuracy in this example is 0.4. Even though CVAE++ generates relevant articles as well, some irrelevant articles could be recommended as well such as the recommended article \#7 entitled "\textit{Optimizing search engines using clickthrough data}", which is more about search engines than RSs. After we examine multiple examples, we can conclude that our model identifies the users' preferences more accurately, especially in the presence of limited data.  

	\subsubsection{\textbf{RQ2}}
\begin{figure}
		\centering
		\subfloat[Citeulike-a, P=1]{\includegraphics[width=0.23\textwidth]{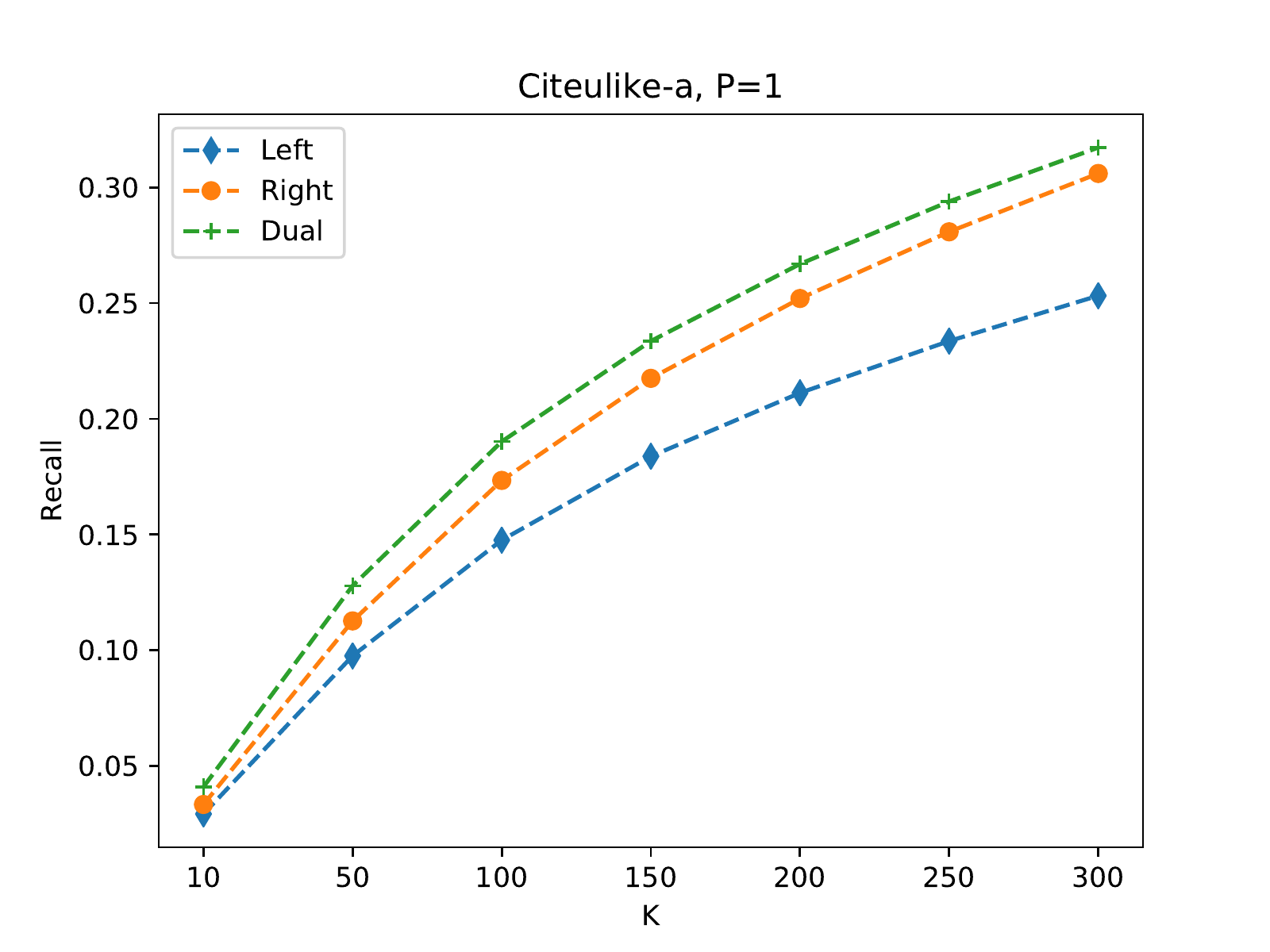} \label{fig:A1}}
		\hfill
		\subfloat[Citeulike-a, P=10]{\includegraphics[width=0.23\textwidth]{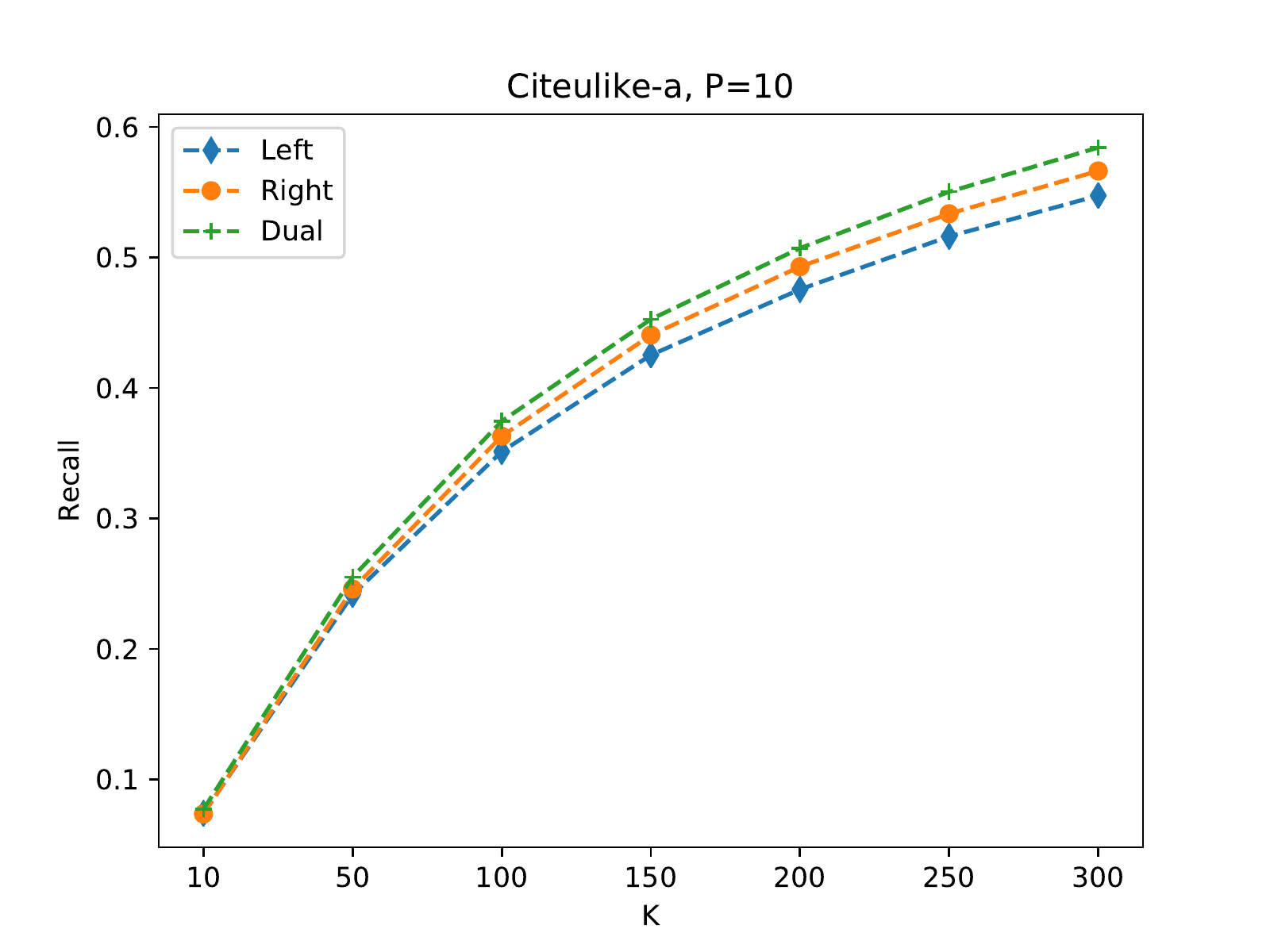} \label{fig:A10}}
		\hfill
		\subfloat[Citeulike-t, P=1]{\includegraphics[width=0.23\textwidth]{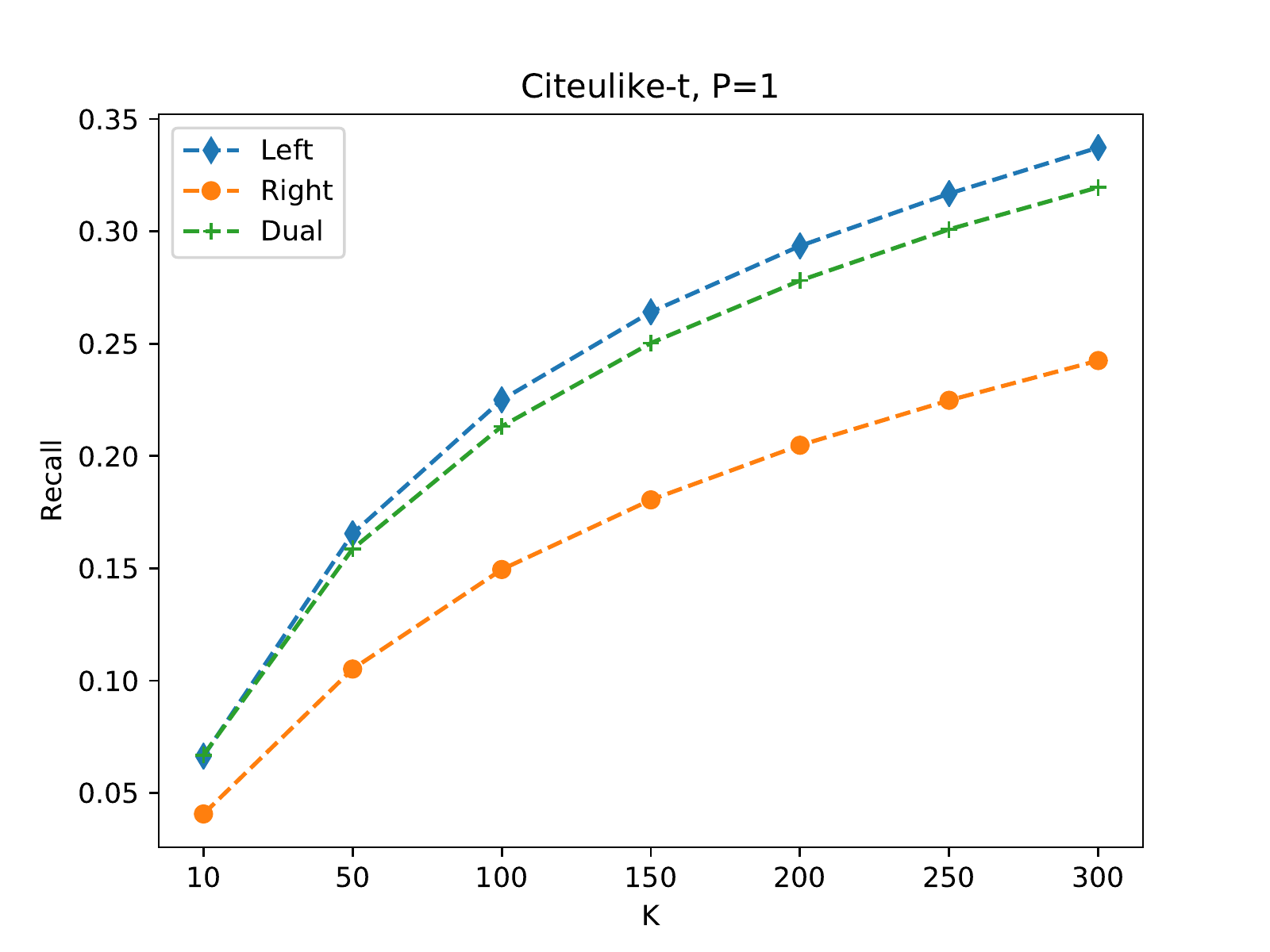} \label{fig:T1}}
		\hfill
		\subfloat[Citeulike-t, P=10]{\includegraphics[width=0.23\textwidth]{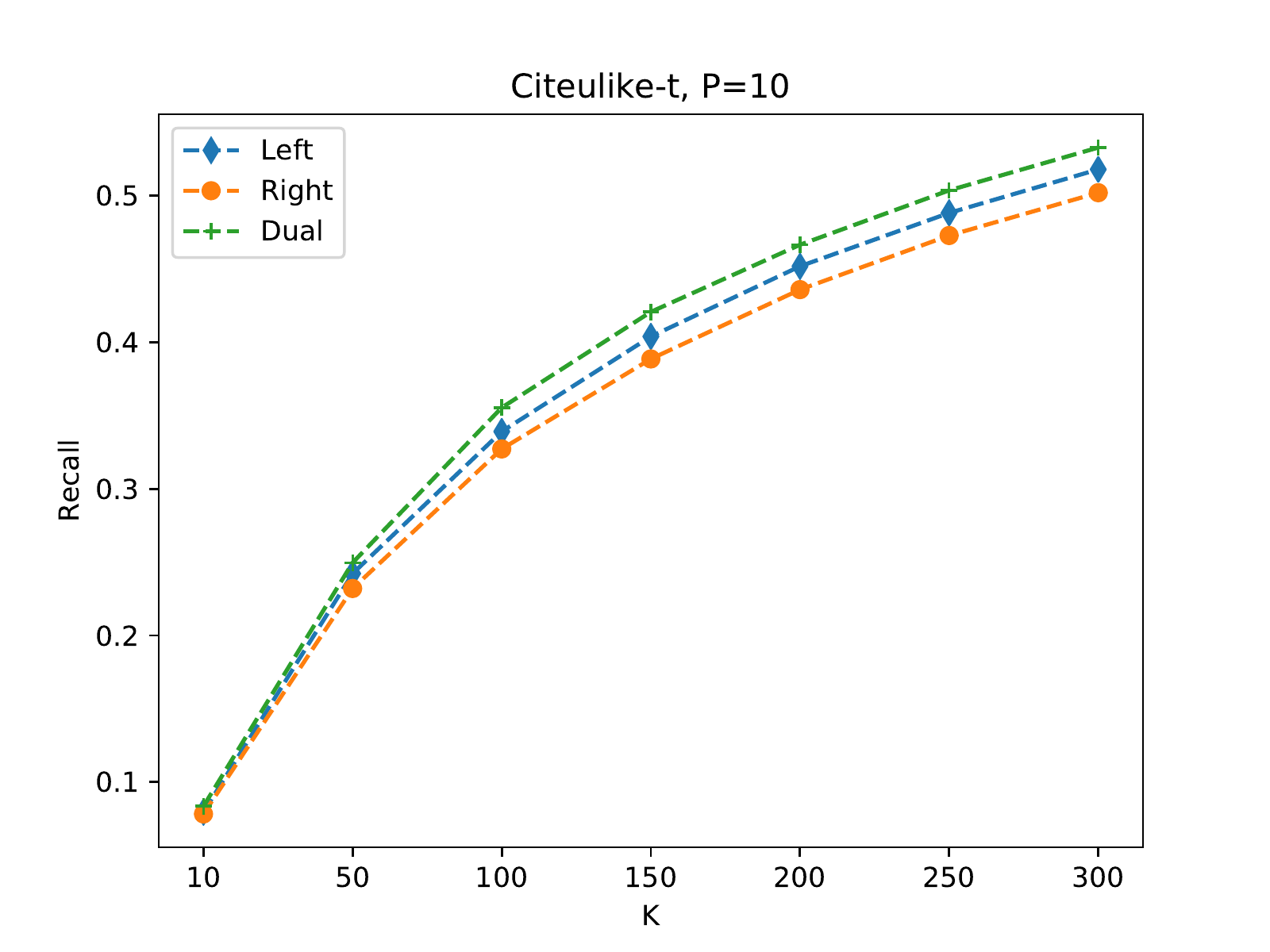} \label{fig:T10}}
		\hfill
		\subfloat[Citeulike-2004-2007, P=1]{\includegraphics[width=0.23\textwidth]{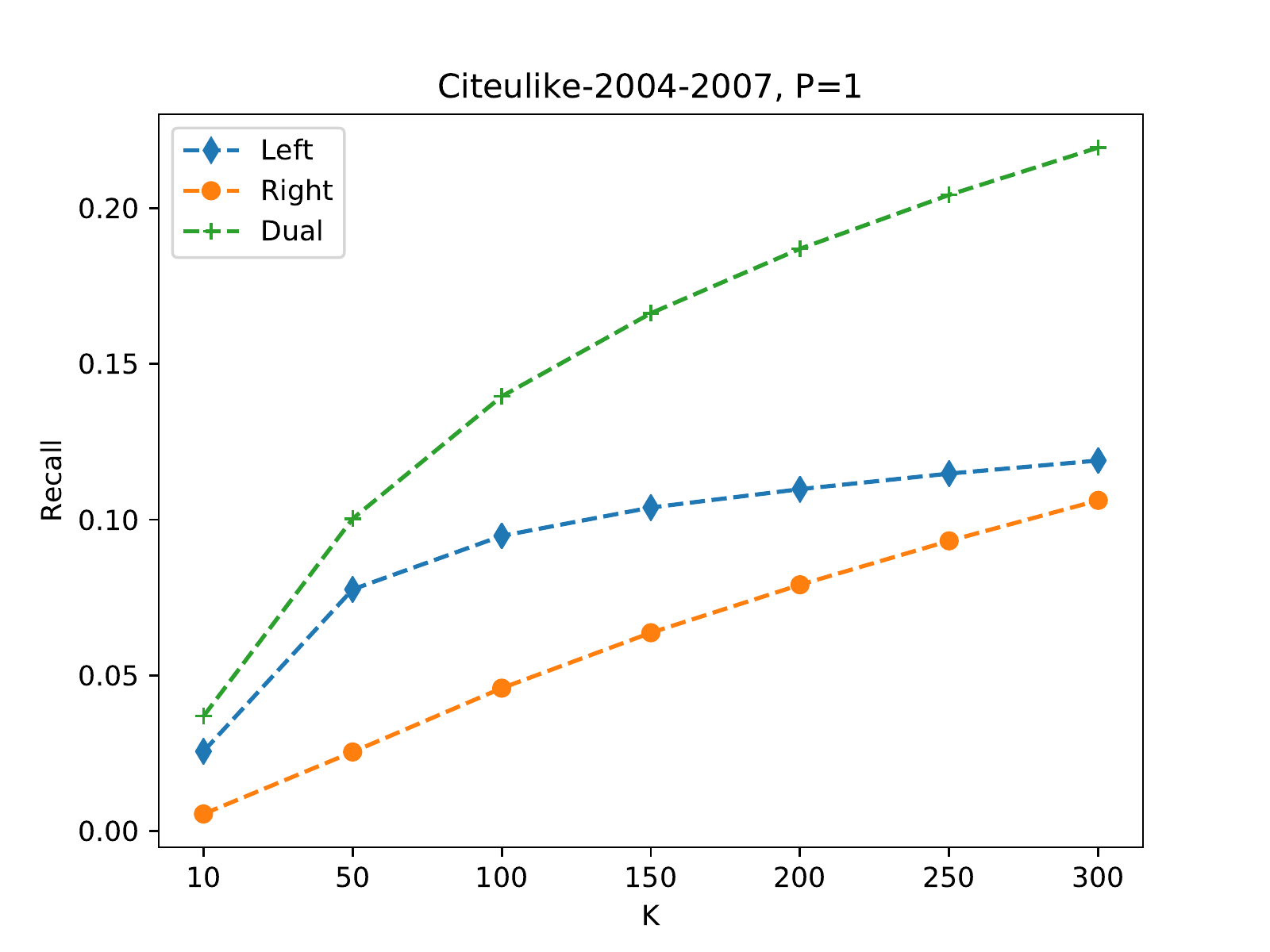} \label{fig:71}}
		\hfill
		\subfloat[Citeulike-2004-2007, P=10]{\includegraphics[width=0.23\textwidth]{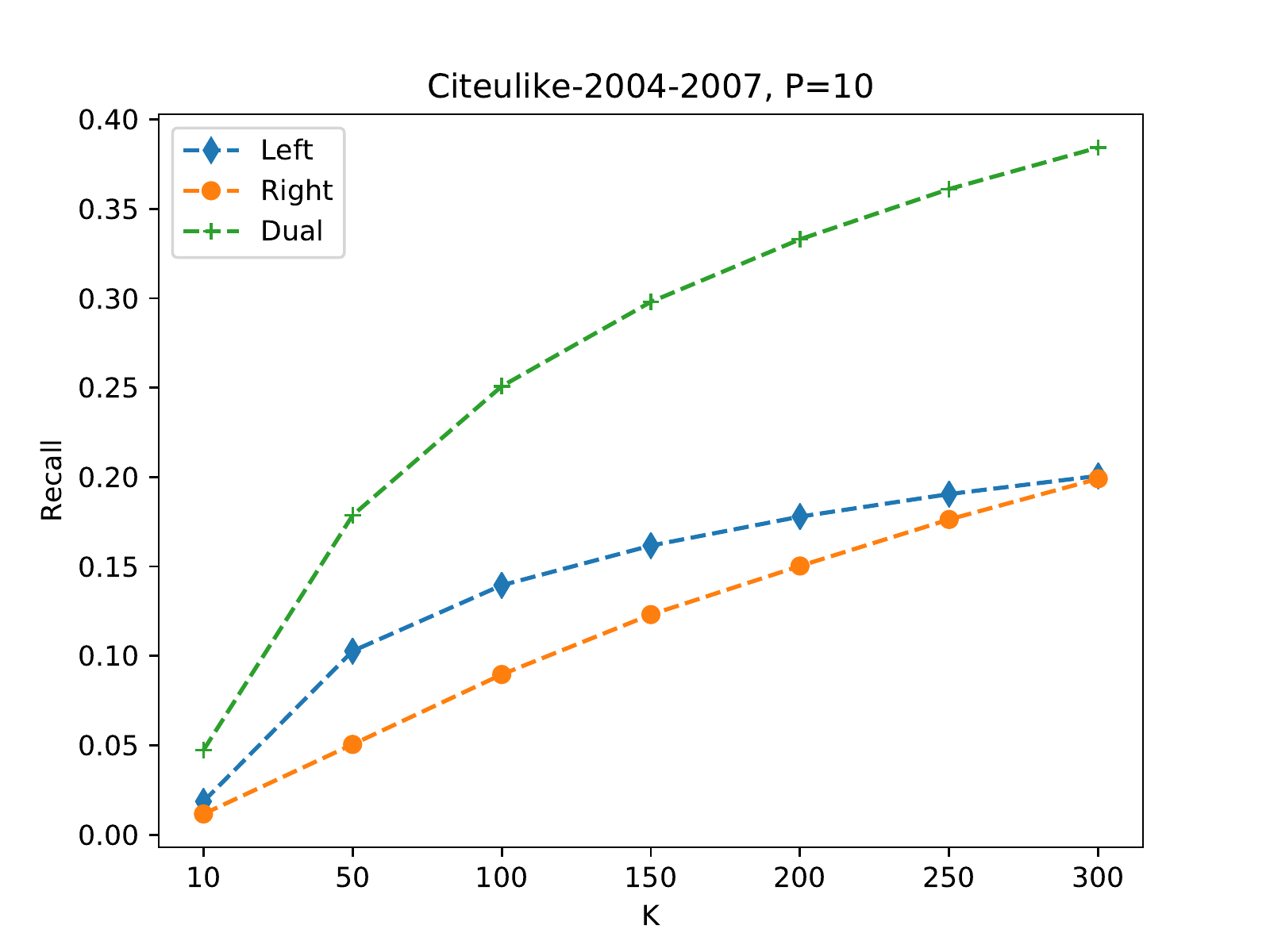} \label{fig:710}}
		\caption{The performance results of using left autoencoder vs. right autoencoder compared to the use of both autoencoders all together for (a-b) Citeulike-a, (c-d) Citeulike-t, and (e-f) Citeulike-2004-2007 datasets.}
		\label{fig:leftandright}
	\end{figure}

	To examine if the two autoencoders are cooperating with each other in finding more similarities between users and items, we run multiple experiments to show how each autoencoder performs solely compared to how they perform all together. In other words, we compare the performance of using the both autoencoders all together in a parallel way (i.e., CATA++) against the performance of using only the right autoencoder (i.e., CATA) that leverages the articles' titles and abstracts, and against the performance of using only the left autoencoder that leverages the articles' tags and citations if available. Figure \ref{fig:leftandright} shows the overall results. As the figure shows, the dual-way strategy has always better results than using each autoencoder solely except of one case in Figure \ref{fig:T1}. In addition, the performance of the left autoencoder and the right autoencoder are competitive to each other, such that the right autoencoder is better than the left autoencoder in Citeulike-a dataset, while the left autoencoder is better than the right autoencoder in the other two datasets. We can conclude that our model by coupling both autoencoders all together is able to identify more similarities between users and items, which leads eventually to better recommendations.      
	
	\subsubsection{\textbf{RQ3}}
	\begin{figure}
		\centering
		\subfloat[Latent space dimension]{\includegraphics[width=0.23\textwidth]{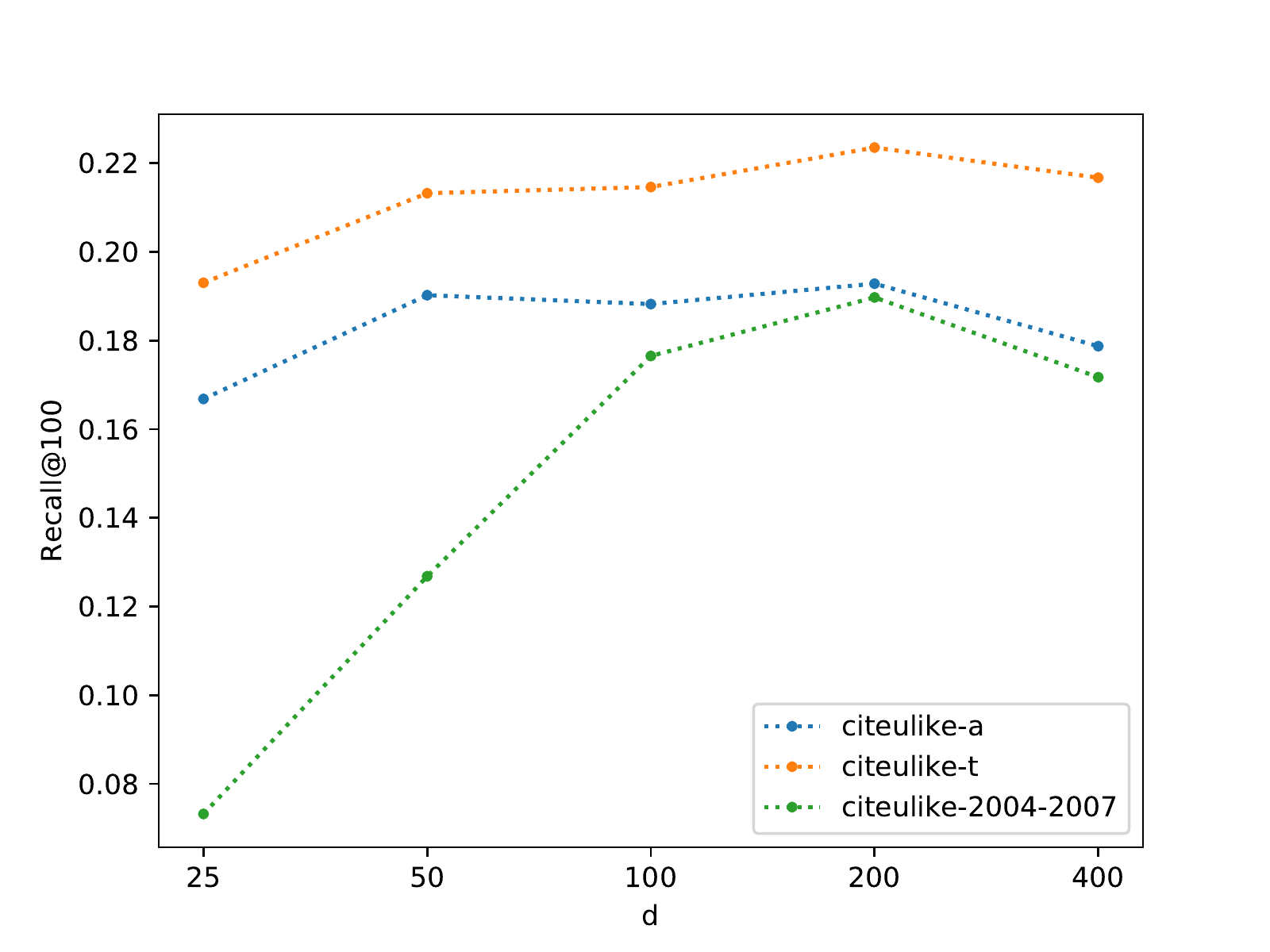} \label{fig:latentSpace}}
		\hfill
		\subfloat[Number of layers]{\includegraphics[width=0.23\textwidth]{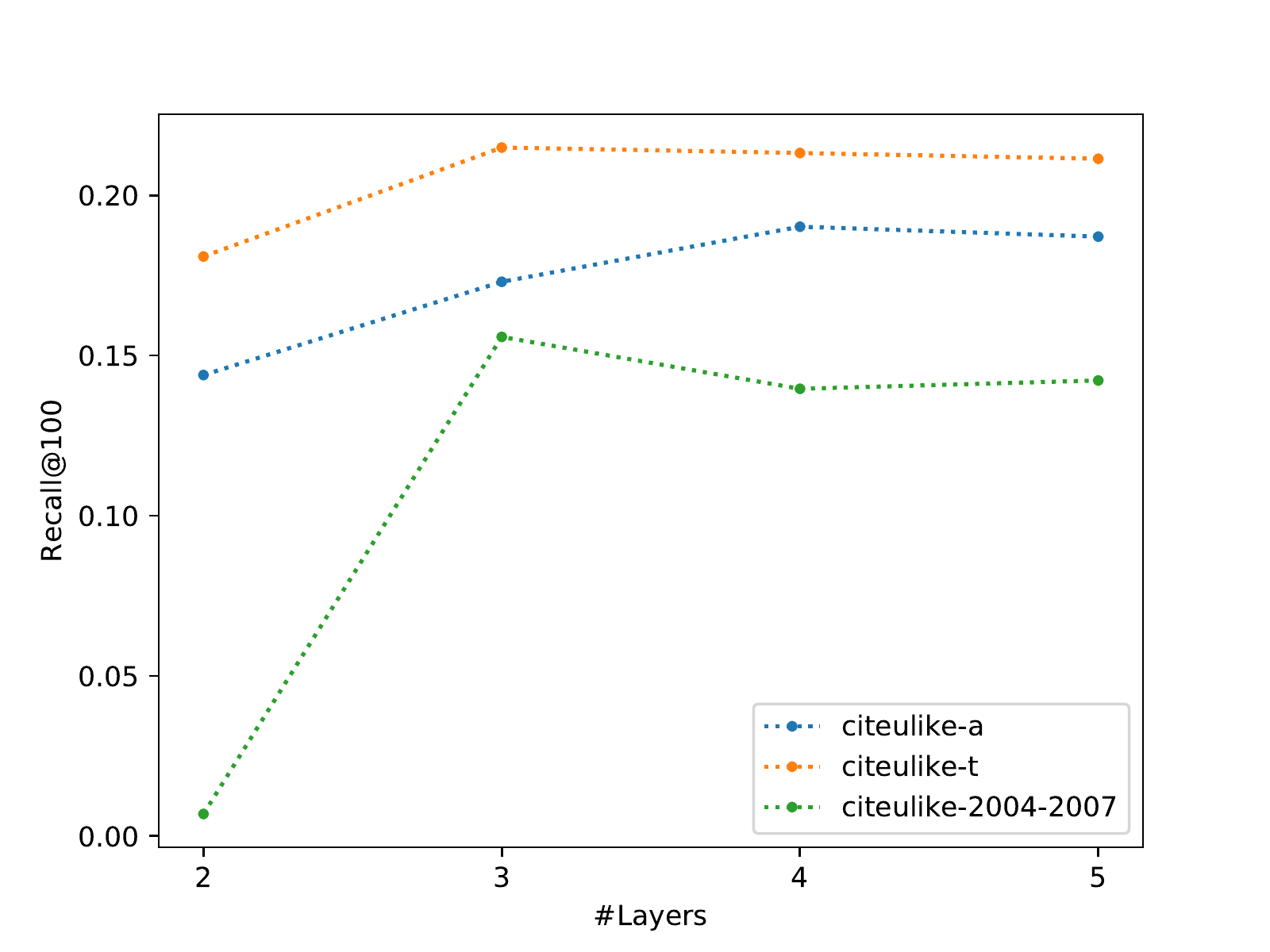} \label{fig:hiddenlayers}}
		\caption{The impact of hyper-parameters tuning on CATA++ performance for: (a) dimension of features' latent space, and (b) number of layers inside each encoder and decoder.}
		\label{fig:hyperparameters}
	\end{figure}
	We conduct several experiments to find out the influence of tuning some hyper-parameters on the performance of our model, such as the dimension of the latent features, the number of hidden layers of the attentive autoencoder, and the two regularization parameters, $\lambda_u$ and $\lambda_v$, used to learn the user/article latent features.
	
	First, the dimension of the latent space used to report our results in the previous section is 50, i.e., each user and item latent feature, $u_i$ and $v_j$, is a vector of size 50. We use the exact number as the state-of-the-art approach, CVAE, in order to have fair comparisons. However, to see the impact of different dimension sizes, we repeat our whole experiments by changing the size into one of following values \{25, 50, 100, 200, 400\}. In other words, we set the size of the latent factors of PMF and the size of the bottleneck of the attentive autoencoder to one of these values. As a result, on average, we observe that when the dimension size is equal to 200, our model has the best performance among all three datasets as Figure \ref{fig:latentSpace} shows. Generally, setting the latent space with size between 100 and 200 is enough to have a reasonable performance compared to the other values.
	
	Second, a four-layer network is used to construct our AAE when we report our results previously. The four-layer network has a shape of "\#Vocabularies-400-200-100-50-100-200-400-\#Vocabularies". However, we again repeat the whole experiments with different number of layers starting from two to five layers, such that each layer has a half size of the previous one. As Figure \ref{fig:hiddenlayers} shows, using less than three layers are not enough to learn the side information. Generally, three-layer and four-layer networks are good enough to train our model.
	
	\begin{figure}
		\centering
		\subfloat[Citeulike-a, P=1]{\includegraphics[width=0.23\textwidth]{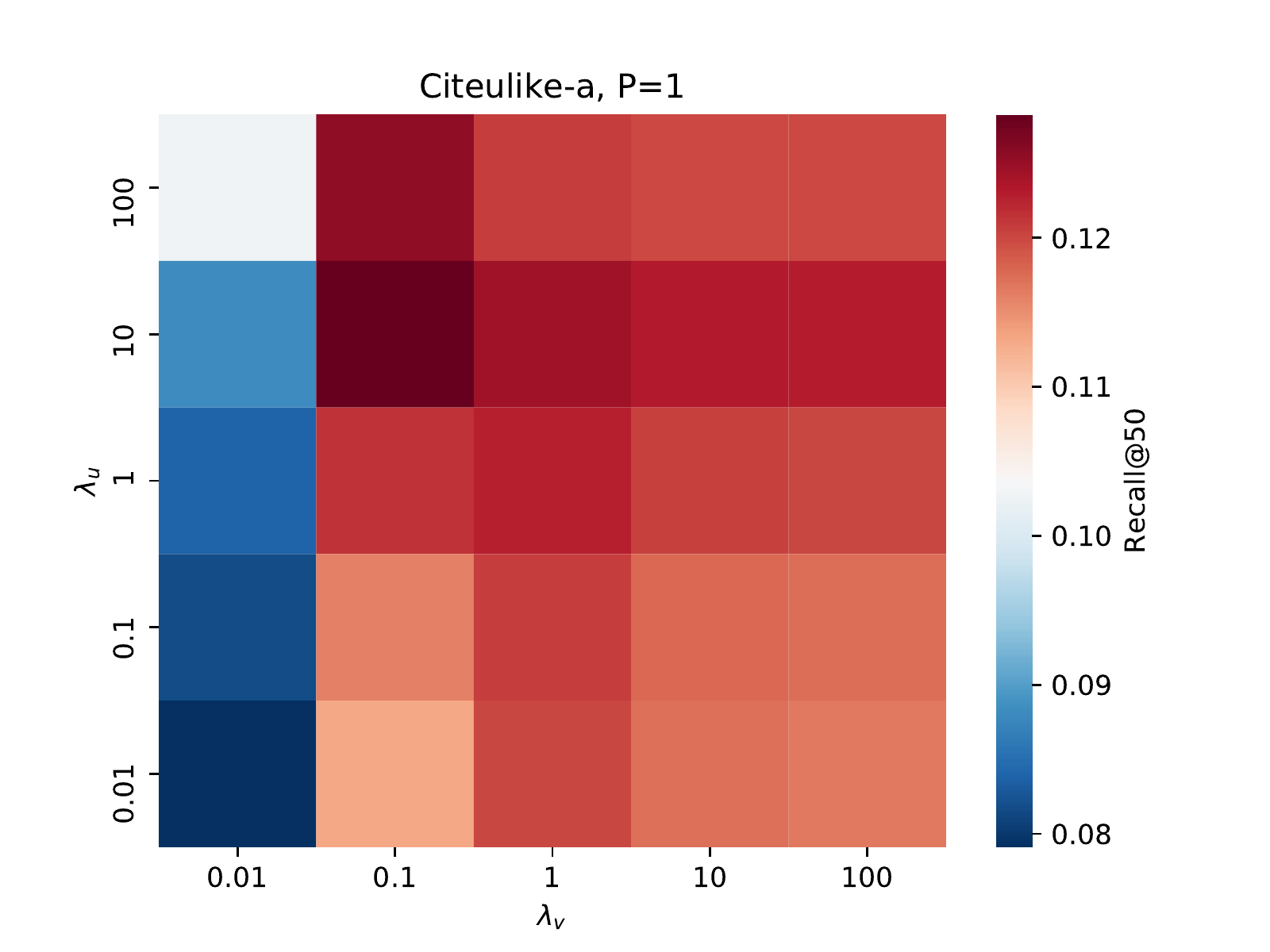} \label{fig:A-1}}
		\hfill
		\subfloat[Citeulike-a, P=10]{\includegraphics[width=0.23\textwidth]{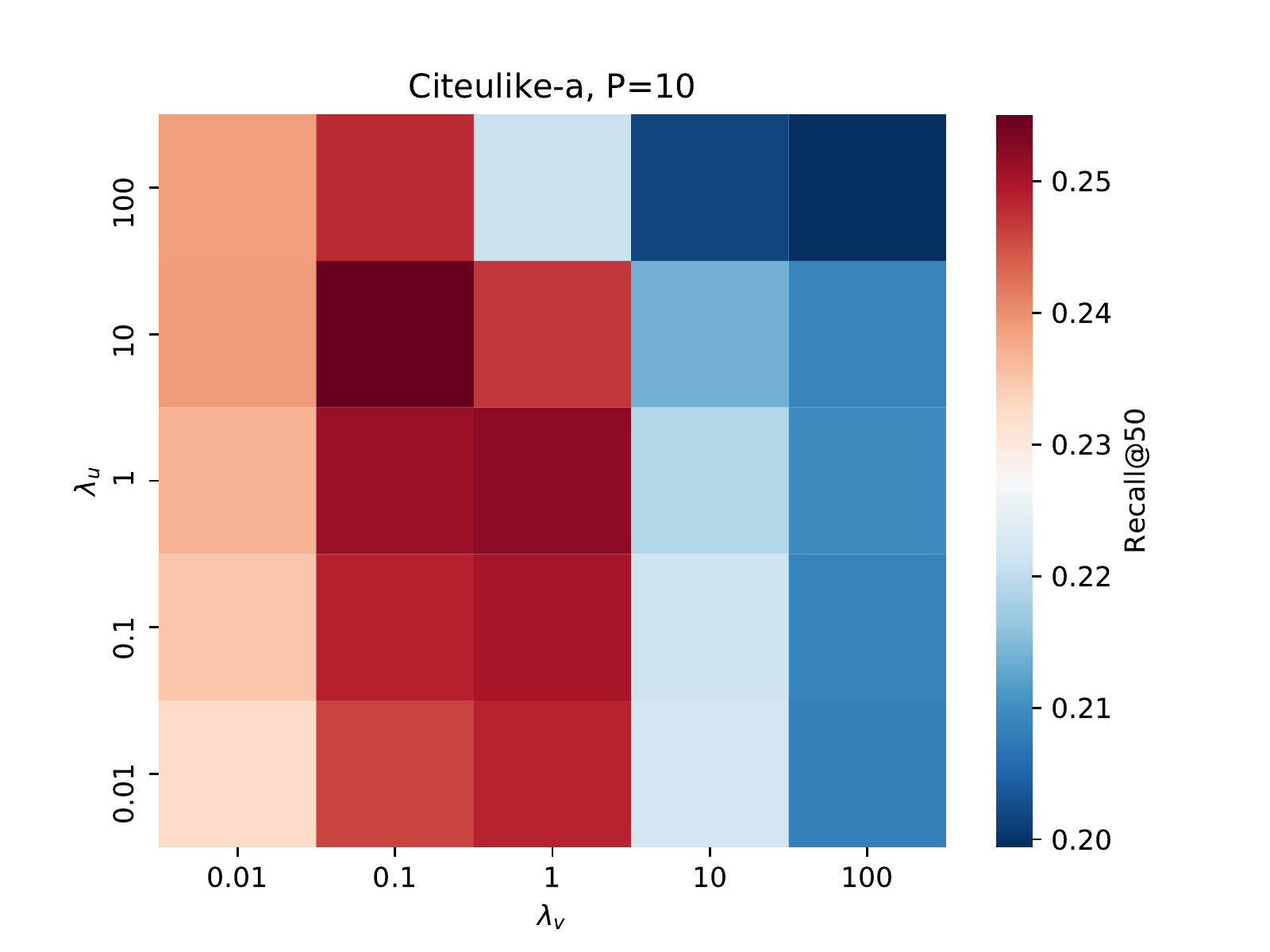} \label{fig:A-10}}
		\hfill
		\subfloat[Citeulike-t, P=1]{\includegraphics[width=0.23\textwidth]{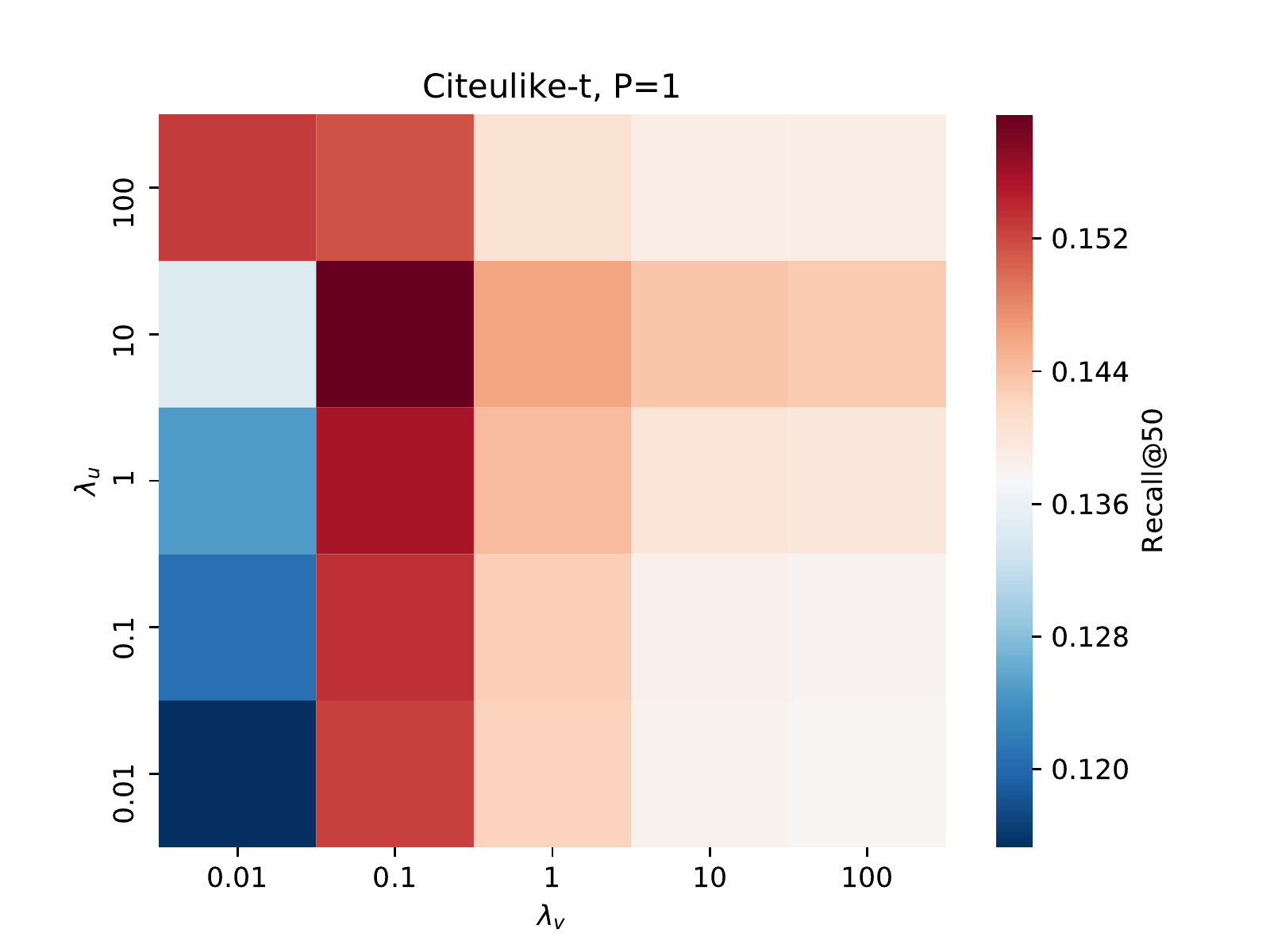} \label{fig:T-1}}
		\hfill
		\subfloat[Citeulike-t, P=10]{\includegraphics[width=0.23\textwidth]{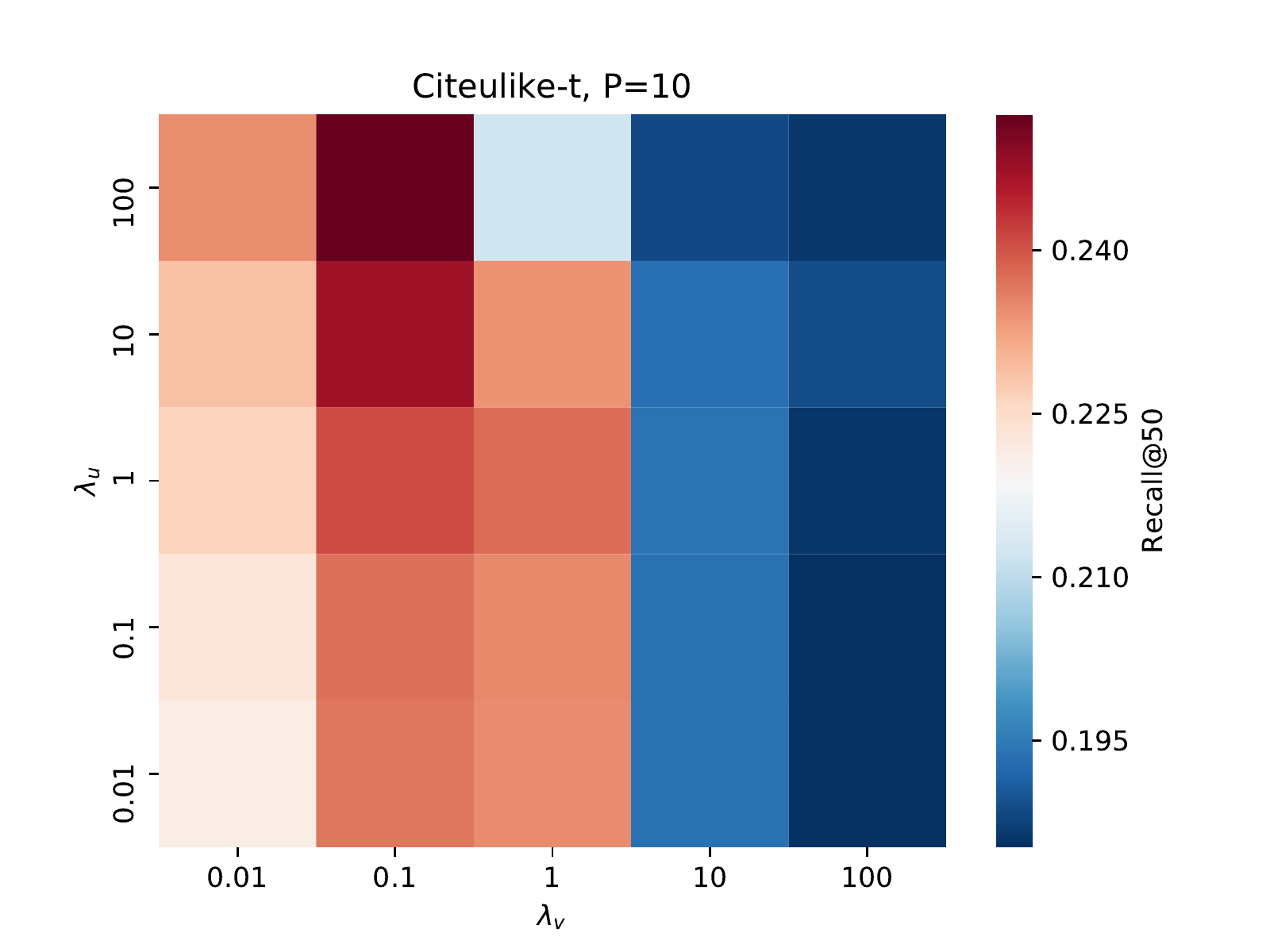} \label{fig:T-10}}
		\hfill
		\subfloat[Citeulike-2004-2007, P=1]{\includegraphics[width=0.23\textwidth]{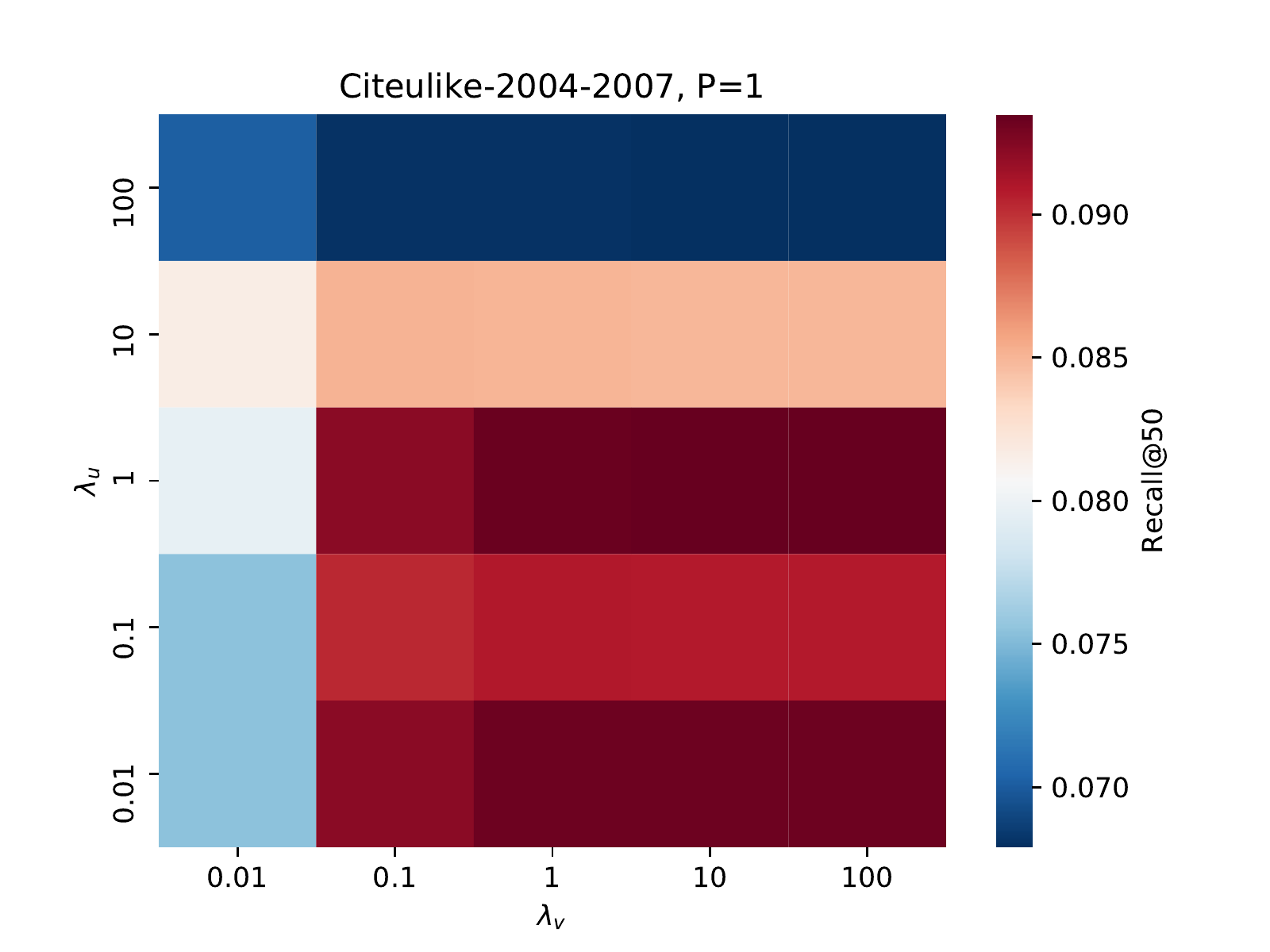} \label{fig:7-1}}
		\hfill
		\subfloat[Citeulike-2004-2007, P=10]{\includegraphics[width=0.23\textwidth]{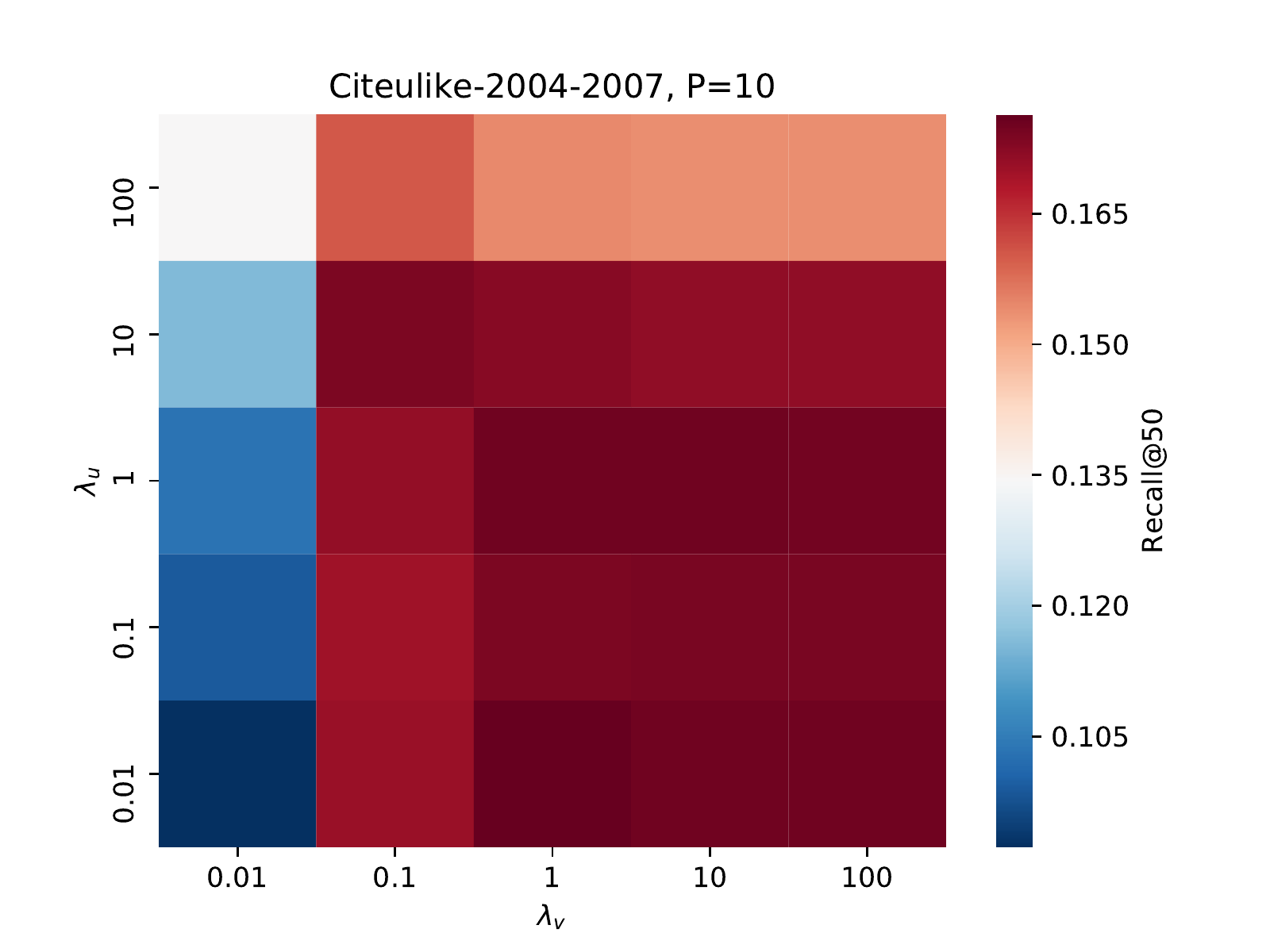} \label{fig:7-10}}
		\caption{The impact of $\lambda_u$ and $\lambda_v$ on CATA++ performance for (a-b) Citeulike-a, (c-d) Citeulike-t, and (e-f) Citeulike-2004-2007 datasets.}
		\label{fig:lambda}
	\end{figure}
	Third, we repeat the experiment again with different values of $\lambda_u$ and $\lambda_v$ from the following range \{0.01, 0.1, 1, 10, 100\}. Figures \ref{fig:A-1} and \ref{fig:T-1} show the performance for the sparse data of Citeulike-a and Citeulike-t datasets, respectively. From these two figures, using a lower value of $\lambda_v$ typically results in lower performance, meaning the user feedback data is not enough and the model needs more article information. The same thing can be said to both scenarios of Citeulike-2004-2007 dataset in Figures \ref{fig:7-1} and \ref{fig:7-10}. In Figure \ref{fig:7-1}, higher value of $\lambda_u$ decrease the performance where user feedback is scarce. Even though Figure \ref{fig:7-10} shows the performance under the dense setting for Citeulike-2004-2007 dataset, it still exemplifies the sparsity with regard to articles as we indicate before in Figure \ref{fig:ratio_items}, where 80\% of the articles are only added to one user library. On the other hand where user feedback is considerably enough, higher value of $\lambda_v$ results in lower performance as Figures \ref{fig:A-10} and \ref{fig:T-10} show.

	\section{Conclusion}
	In this paper, we alleviate the natural data sparsity problem in recommender systems by introducing a dual-way strategy to learn item's textual information by coupling two parallel attentive autoencoders together. The learned item's features are then utilized in the learning process of matrix factorization (MF). We evaluate our model for academic article recommendation task using three real-world datasets. The huge gap in the experimental results validates the usefulness of exploiting more item's information, and the benefit of integrating attention technique in finding more relevant recommendations, and thus boosting the recommendation accuracy. As a result, our model, CATA++, has the superiority over multiple state-of-the-art MF-based methods according to several evaluation measurements. Furthermore, the performance of CATA++ is improved the most where the data sparsity is the highest.
	
	For future work, new metric learning algorithms could be explored to substitute MF technique because the dot product in MF doesn't guarantee the triangle inequality \cite{CML}. For any three items, the triangle inequality is fulfilled once the sum of distance between any two item pairs in the feature space should be greater or equal to the distance of the third item pair, such that $d(x,y) \leq d(x,z) + d(z,y)$. By doing so, user-user and item-item relationships might be captured more accurately. 
	
	\printcredits
	
	\bibliographystyle{cas-model2-names}
	
	\bibliography{CATA++}

	%
\end{document}